\documentclass[10pt,twocolumn,letterpaper]{article}

\usepackage{cvpr}              
\usepackage{booktabs}
\usepackage{arydshln} 
\usepackage[accsupp]{axessibility}

\usepackage{booktabs}

\definecolor{cvprblue}{rgb}{0.21,0.49,0.74}
\usepackage[pagebackref,breaklinks,colorlinks,allcolors=cvprblue]{hyperref}

\def\paperID{42408} 

\def\confName{CVPR}
\def\confYear{2026}

\newcommand{\ourmethod}{\textit{TokenLight}}

\title{\ourmethod: Precise Lighting Control in Images using Attribute Tokens    }

\author{Sumit Chaturvedi$^{1 *}$ \quad Yannick Hold-Geoffroy$^2$ \quad Mengwei Ren$^2$ \quad Jingyuan Liu$^2$ \\ He Zhang$^2$ \quad Yiqun Mei$^2$ \quad Julie Dorsey$^1$ \quad Zhixin Shu$^{2 \dag}$ \vspace{0.2cm}\\
{\hspace{-17mm} $^1$ Yale University} \quad {\hspace{-1.5mm}$^2$ Adobe} }
\newcommand\blfootnote[1]{%
  \begingroup
  \renewcommand\thefootnote{}\footnote{#1}%
  \addtocounter{footnote}{-1}%
  \endgroup
}

\begin{document}

\twocolumn[{%
\renewcommand\twocolumn[1][]{#1}%
\maketitle 
\begin{center} 
\vspace{-0.23in}
\centering 
\includegraphics[width=\linewidth]{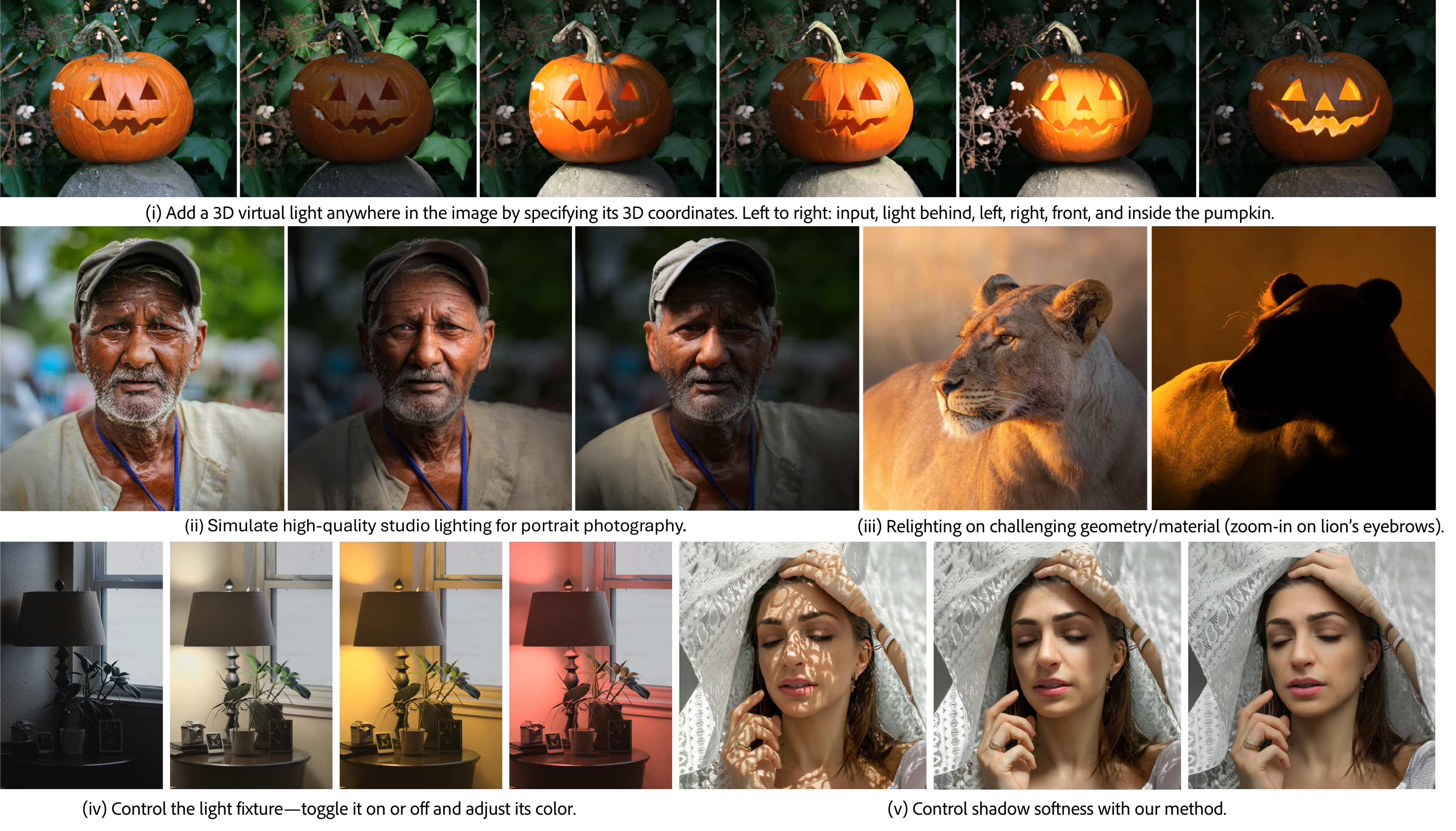}
    \captionof{figure}
    {\ourmethod\ enables intuitive, flexible lighting edits for any image. Using a tokenized lighting-attribute design for diffusion transformers, it provides precise control over a light's intensity, color, diffuseness, 3D location, and more. This enables effects like placing a light inside a pumpkin, adding back-lighting to a lion, simulating Rembrandt-style setups, diffusing shadows and shifting color tones.}
    \label{fig:teaser}
\end{center}
}]

\blfootnote{*Work done as an intern at Adobe Research.}
\blfootnote{\dag Corresponding author.}

\begin{abstract}
This paper presents a method for image relighting that enables precise and continuous control over multiple illumination attributes in a photograph. We formulate relighting as a conditional image generation task and introduce attribute tokens to encode distinct lighting factors such as intensity, color, ambient illumination, diffuse level, and 3D light positions. The model is trained on a large-scale synthetic dataset with ground-truth lighting annotations, supplemented by a small set of real captures to enhance realism and generalization. We validate our approach across a variety of relighting tasks, including controlling in-scene lighting fixtures and editing environment illumination using virtual light sources, on synthetic and real images. Our method achieves state-of-the-art quantitative and qualitative performance compared to prior work. Remarkably, without explicit inverse rendering supervision, the model exhibits an inherent understanding of how light interacts with scene geometry, occlusion, and materials, yielding convincing lighting effects even in traditionally challenging scenarios such as placing lights within objects or relighting transparent materials plausibly. Project page: \url{vrroom.github.io/tokenlight/}
\end{abstract} 
\section{Introduction}

Lighting control in images is a fundamental task in modern creative workflows, with broad applications in photography, post-production, visual effects, and mixed reality  \cite{kim2024switchlight, inverse_lighting_and_ar_2010}. The ability to precisely manipulate illumination not only enhances aesthetic quality but also ensures visual consistency across scenes, objects, and composite elements. Traditionally, lighting is configured either through physical lighting equipment or within 3D rendering software, where such control becomes inaccessible once the image has been captured or the scene has been rendered. 

Recent advances in image generation have enabled image relighting in various forms. The utility of these approaches is largely determined by how lighting is represented---namely, how the desired illumination conditions are encoded and controlled. Different representations provide different levels of precision and flexibility in interacting with the scene in an image. Notably, text-driven relighting \cite{text_2_relight_aaai_25, iclight_iclr_2025} provides intuitive, language-based control, but relies on the model’s interpretation of user prompts, often resulting in imprecise or unpredictable outcomes. Background-image--based relighting \cite{iclight_iclr_2025, ren2024relightful} offers limited information about the underlying illumination, restricting fine-grained control. Panoramic environment maps \cite{jin2024neural_gaffer, kim2024switchlight, diffusion_renderer_cvpr_25, pandey2021total, chaturvedi2025synthlight, zeng2024dilightnet}, while capturing lighting from all directions, cannot model near-field or spatially localized illumination changes. Finally, inverse-rendering--based methods \cite{ponglertnapakorn2023difareli, cai_2024_cvpr, zeng2024dilightnet, zhang2025relitlrm} recover scene geometry and material properties to enable lighting modification, but achieving high-quality relighting demands accurate 3D reconstruction, which remains challenging, especially in single input view settings. Moreover, 2.5D approaches \cite{physically_controllable_relighting_sig_25, pandey2021total, kim2024switchlight, erel2025practilightpracticallightcontrol} that estimate surface normals or depth maps provide only partial geometric cues and often fail to handle occlusions or light interactions across hidden surfaces. What is missing is a representation that allows precise, interpretable, and spatially localized adjustments to illumination directly in the image domain---bridging the intuitive flexibility of 3D lighting tools with the accessibility of 2D image editing.

To this end, we introduce \ourmethod, a conditional image generation framework that employs a tokenized lighting representation, where physically meaningful tokens encode illumination attributes, to model and control distinct lighting factors. Each attribute token corresponds to a light property such as intensity, color, diffuse level, or 3D spatial position, enabling continuous and incremental manipulation of lighting effects in a disentangled manner. This formulation allows \ourmethod\ to produce realistic, geometry-aware relighting results without requiring 3D reconstruction, offering a new level of precision and interpretability in image-based lighting control. Furthermore, it naturally extends to in-scene lighting control, such as turning individual light sources on or off within an image, by combining spatial masks for target light sources with attribute tokens specifying the desired illumination characteristics.

Achieving physically meaningful control in relighting requires the model to understand how image appearance changes in response to precise combinations of lighting attributes. Consequently, model training demands accurate annotations of lighting variations, which are difficult to obtain in real-world settings but can be readily simulated in a modern 3D rendering software like Blender \cite{blender}. We therefore synthesize a large-scale dataset of 3D assets rendered under systematically varied lighting conditions using a path-tracing renderer, providing precise ground-truth supervision for each lighting attribute. With this dataset, we formulate relighting as an end-to-end conditional generation process, taking control signals as input and directly re-rendering the desired output using a diffusion-based model.

To build upon a strong visual prior, we leverage a foundation diffusion transformer pretrained for text-to-image and text-to-video generation, and fine-tune it on our curated dataset for controllable relighting. In addition, we incorporate a small set of real-world captures in which lighting changes correspond to toggling in-scene light sources, further improving realism and generalization.

We present three forms of practical lighting controls to demonstrate the strength and flexibility of our framework:
(1) adding spatial virtual lights,
(2) editing or diffusing environment illumination, and
(3) controlling in-scene light sources.
We show that different combinations of attribute tokens enable rich and creative lighting manipulations. Our method is validated both qualitatively and quantitatively on real-world and synthetic datasets. Results demonstrate that our end-to-end control-to-render approach, without requiring explicit inverse graphics of the input image, exhibits strong reasoning ability and a robust understanding of light–scene interaction learned purely from synthetic data. For instance, one can use the model to place virtual lights behind objects (e.g., \cref{fig:careaga_comparison}(i)) or inside a pumpkin to create a glowing jack-o’-lantern (see \cref{fig:teaser}), or add lights near transparent surfaces to produce plausible shadows (see \cref{fig:spatial_results}).

%

We summarize our contributions as follows:

\begin{itemize}
\item Formulating a precise and continuous lighting control in images as an end-to-end image generation problem, where lighting attributes are combined with an input image to directly re-render the desired output using a diffusion-based model.
\item Introducing a compact, physically meaningful lighting-attribute token representation that unifies various relighting tasks within a diffusion transformer architecture.
\item Achieving state-of-the-art performance across multiple relighting tasks, showcasing the model’s reasoning ability and robust understanding of light–scene interaction learned from synthetic supervision.
\end{itemize}

\section{Related Work}
\label{sec:related_work}

\begin{figure*}[!htbp]
    \centering
    \includegraphics[width=\textwidth]{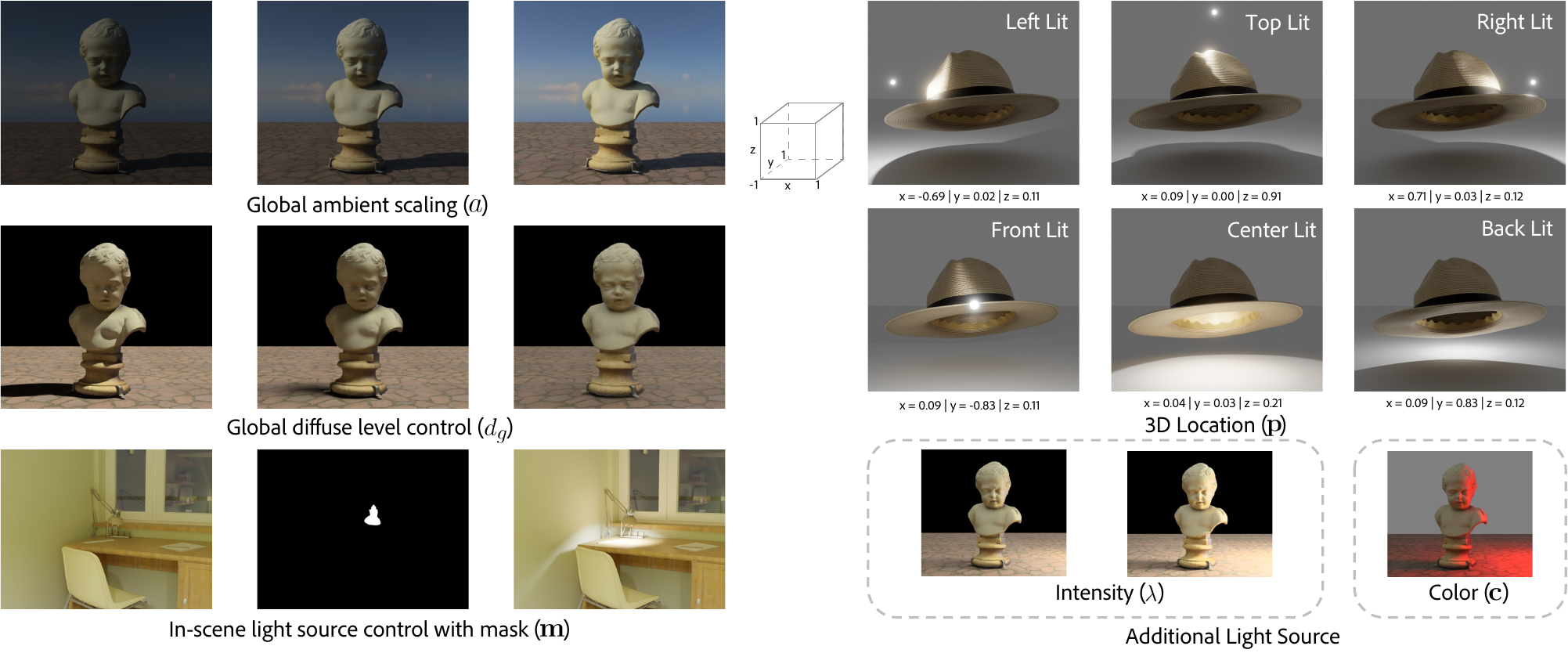}
    \vspace{-8mm}
    \caption{Overview of our data construction and lighting controls.
We generate per-light renders, ambient images, and light-visibility masks to supervise the model. The system supports ambient scaling, global diffuse-level control, and masks for visible in-scene lights, and enables adding new lights at 3D positions with specified color and intensity. The bright light markers are visualization only.}
    \label{fig:synthetic_data}
\end{figure*}

\subsection{Relighting}


Relighting seeks to modify the illumination in images or videos while maintaining the original content's appearance and identity. Early approaches relied on image-based techniques, where a reflectance field is constructed by capturing multiple images of a subject under varying illuminations. These captures are then used to synthesize relit images \cite{debevec_acquiring_reflectance_field_2000, ren_image_based_relighting_tog_2015,xu2018deep}. 
Later, methods adopted an inverse-rendering framework \cite{barron2014shape, nerd_boss_iccv_2021, dib_r_pp_chen_neurips_2021, diffusion_renderer_cvpr_25, nvdiffrecmc_hasselgren_neurips_2022, gaussian_shader_cvpr_2024, ws_sir_yi_cvpr_2023}, where scene properties such as geometry, materials, and lighting are estimated to enable re-rendering under novel illumination conditions. For a comprehensive review of these techniques, please refer to the survey in \cite{neural_relighting_survey_egsr_2021}. 
With the rise of generative imaging, especially diffusion models \cite{saharia2022palette,ho2020denoising,Peebles_2023_ICCV,rombach2021highresolution}, photorealistic image generation has become a practical reality. This progress has catalyzed a growing body of work that leverages natural image priors for relighting across diverse domains: objects \cite{jin2024neural_gaffer, zeng2024dilightnet, genlit_sig_asia_shrisha_2025}, portraits \cite{iclight_iclr_2025, ponglertnapakorn2023difareli, ren2024relightful, chaturvedi2025synthlight, lux_post_facto_cvpr_2025}, indoor scenes \cite{light_lab_sig_25, scribble_light_choi_2025_cvpr, luminet_cvpr_2025, stylitgan_bhattad_cvpr_2024, stylelight_wang_eccv_2022} and outdoor environments \cite{kocsis2024lightit, he2025unirelightlearningjointdecomposition, griffiths2022outcast, diffusion_renderer_cvpr_25}. Some of these methods still rely on explicit scene property decomposition to enable user control. However, such intrinsic representations are often difficult to estimate reliably and unintuitive for end users to edit. In particular, accurately representing and editing precise light positions and near-field lighting effects remain a challenge within these frameworks. 

GenLit \cite{genlit_sig_asia_shrisha_2025} addresses near-field effects through a view-agnostic lighting parameterization, which complicates local edits and can lead to incoherent lighting across regions of the image. In contrast, our approach introduces a unified token representation that directly couples lighting with the camera viewpoint. This compact and expressive representation facilitates precise illumination editing, including view-tied placement of the light, and direct adjustment of chromaticity and intensity for visible emitters. Our representation can also handle all three common settings---indoor, outdoor, and object-centric---relatively well, as demonstrated in our experiments. 

\subsection{Light Representations}


One key factor affecting relighting usability is the chosen lighting representation. Some approaches perform intrinsic image decomposition followed by re-synthesis under new lighting \cite{rgbx_sig_24, lyu2025intrinsicedit,diffusion_renderer_cvpr_25}, typically using an environment map \cite{kim2024switchlight, pandey2021total, zeng2024dilightnet, jin2024neural_gaffer, diffusion_renderer_cvpr_25, lux_post_facto_cvpr_2025, comprehensive_relighting_cvpr_2025}. While these methods preserve identity and enable detailed edits, manipulating HDR irradiance or environment maps is laborious and unintuitive for non-expert users. To simplify light editing, methods based on text prompts \cite{iclight_iclr_2025, text_2_relight_aaai_25}, image harmonization \cite{ren2024relightful, zhang2025zerocomp}, or latent space traversal \cite{stylitgan_bhattad_cvpr_2024} offer greater usability but sacrifice fine-grained control, limiting precision in tasks such as adjusting cast shadow directions. Another strategy involves cloning reference lighting from an exemplar \cite{ponglertnapakorn2023difareli, luminet_cvpr_2025, shu_2017_portraitlighting}, but requires an appropriate illumination reference, which is not always feasible. Simple user edits, such as scribbles for highlights and shadow \cite{painting_with_light_sig_93, scribble_light_choi_2025_cvpr, light_painter_cvpr_2023} or direct shadow controls \cite{futschik_cvpr_2023_controllable_light_diffusion, portrait_shadow_manip_zhang_sig_20, generative_shadow_removal_sig_a_2024, face_relit_shadow_cvpr_22, blind_removal_of_facial_foreign_shadows_hou_bmcv_2022, compose_andrew_hou_eccv_2024, fortier2024spotlight, erel2025practilightpracticallightcontrol}, have also been explored, but lack a balance between editing simplicity and control effectiveness. 

Closest to our work are LightLab \cite{light_lab_sig_25} and \citet{physically_controllable_relighting_sig_25}, which provide some of the most advanced yet simple editing controls. The former allows toggling on or off visible emitters via 2D masks but cannot insert or anchor lights in 3D, while the latter exposes a 3D interface but remains a two-stage process that depends on explicit 2.5D reconstruction. This requirement makes handling occluded regions or material with complex properties such as skin and hair difficult, despite their neural rendering fixing pass. In contrast, our method preserves key 3D lighting affordances without requiring explicit scene reconstruction. By training end-to-end on synthetic data, supplemented with a small set of real data captured   under controlled lighting changes \cite{light_lab_sig_25,haeberli1992synthetic}, our method circumvents many of the failure modes associated with geometry and material estimation, while retaining user-friendly control and editability. 
\section{Method}
\label{sec:method} 

Given a scene $S$ and lighting $L$, the image is produced by a rendering function $f$: $I = f(S, L)$. Inverse rendering seeks $S = g(I)$. For relighting, an incremental lighting edit $\Delta L$ yields $I_r = f(S, L_r)$ with $L_r = L + \Delta L$. Substituting $S = g(I)$ gives $I_r = f(g(I), L + \Delta L) = R(I, \Delta L)$, where $R$ is a learned relighting operator, a function that only depends on the input image and lighting change. We parameterize $\Delta L$ as:
\begin{itemize}
  \item \textbf{Global ambient scaling:} $L_r = a \cdot L$, where $a$ is a scalar.
  \item \textbf{Global diffuse-level control:} a scalar $d_g$ that adjusts the diffuse level of the existing lighting; different values of $d_g$ yield different shadow softness and highlight roll-off.
  \item \textbf{Additional light source:} $\Delta L_{\text{add}} = (\mathbf{p}, \mathbf{c}, \lambda, d)$, where $\mathbf{p}$ is the light position, $\mathbf{c}$ its color, $\lambda$ its scalar intensity, and $d$ a per-light diffuse-level control.
  \item \textbf{In-scene light sources:} $\Delta L_{\text{in}} = (\mathbf{m}, \mathbf{c}, \lambda, t)$, where $\mathbf{m}$ is a light mask defining in-image fixtures, $\mathbf{c}$ their color, $\lambda$ their scalar intensity, and $t$ a binary flag to indicate the transition, whether the light source is dimming/brightening.
\end{itemize}
In \emph{TokenLight}, $\Delta L$ is represented by a set of tokens that explicitly encode physical attributes (e.g., intensity coefficients and 3D locations), up to an appropriate normalization factor. We model $R$ with a conditional generative distribution $p_{\theta}$ such that $I_r \sim p_{\theta}(\,\cdot \mid I, \Delta L)$, where $p_{\theta}$ is parameterized by a diffusion model (see \cref{subsec:flow_matching_model}).

\subsection{Datasets for Lighting Attributes Variation}
\label{subsec:synthetic_data}

As shown in Fig. \ref{fig:synthetic_data}, we construct a dataset of controlled lighting attribute variations. Each sample is derived from a 3D scene $S$ created in Blender and rendered using the Cycles physically based path tracer. Given a fixed scene $S$ and camera, we render one or more image pairs $\langle I, I_r \rangle$ under different lighting edits $\Delta L$, sampled over combinations of the following attributes: 
(i) global ambient scale $a$, 
(ii) global diffuse level $d_g$, 
(iii) an additional light $\Delta L_{\text{add}} = (\mathbf{p}, \mathbf{c}, \lambda, d)$, and 
(iv) in-scene fixtures $\Delta L_{\text{in}} = (\mathbf{m}, \mathbf{c}, \lambda, t)$.

For global ambient scaling $a$, we modulate the intensity of an HDRi environment map or an ambient-color environment texture.  
For global diffuse control $d_g$, each scene is illuminated by an ambient term and a dominant area light; the diffuse level corresponds to the light’s angular spread, which governs the beam width and thus the softness of shadows---larger spreads yield broader illumination and smoother shading transitions.  
For the additional light term $\Delta L_{\text{add}}$, we add a single point light defined by its position $\mathbf{p}$, color $\mathbf{c}$, intensity $\lambda$, and a radius factor $d$ controlling the falloff and diffusion of light. The light position is specified relative to the camera for consistent editing behavior (see \cref{subsec:representing_3d_light_positions} for details).  
For the in-scene lighting term $\Delta L_{\text{in}}$, we employ an image-plane segmentation mask $\mathbf{m}$ to define target light source regions within the scene, whose intensity $\lambda$ and color $\mathbf{c}$ are parameterized analogously to the added virtual light.  
Together, these configurations yield a diverse and physically grounded dataset of image pairs $(I, I_r)$ with explicit supervision on the lighting attribute change $\Delta L$.

Our data is rendered from a diverse pool of 3D assets. For diffuse control and additional virtual-light edits, we use procedurally generated scenes built from a filtered Objaverse subset \cite{jin2024neural_gaffer}, with optional ground and wall geometry to catch shadows. We also add procedurally generated 3D humans following \cite{wood2021fake, chaturvedi2025synthlight, facelift_lyu_2025_iccv}. During rendering, each scene samples 64 point-light positions and may draw from among $\sim$600 HDRi from PolyHaven \cite{polyhaven}. Diffuse control supervision is obtained with an area light rendered at 6 random spread angles, normalized to $[0,1]$ for training. 

For in-scene lighting, we use 83 artist-authored indoor scenes with annotated visible fixtures and render each light’s contribution separately, producing $\sim$100K images. To complement the synthetic corpus, we capture $\sim$600 real indoor photographs. Rendered images are denoised and stored in linear RGB, producing aligned triplets $(I,\Delta L,I_r)$ that correspond to intended lighting edits; the real photographs provide cross-domain realism and diversity and are used for paired supervision, with unknown lighting change attributes (e.g., color) dropped out.
\paragraph{Paired Supervision Synthesis.}

To avoid exhaustively rendering every lighting variation, we generate supervision pairs on-the-fly by linearly combining pre-rendered lighting components in linear RGB. Each relit image is a weighted sum of an environment-map render (for ambient $a$) and a set of ``on'' light renders for additional-light and in-scene terms with attributes $(\mathbf{p}, \mathbf{m}, d)$. Ambient and light intensities $(a,\lambda)$ and color $\mathbf{c}$ are randomly sampled during data loading. Diffuse control $d_g$ is achieved by mixing light renders with different angular spreads to vary shadow softness. All synthesized images are tone-mapped with the Reinhard operator \cite{reinhard_tone_mapping}. 
Further details appear in the supplementary.

\subsection{Scene-Agnostic Camera and Lighting}
\label{subsec:representing_3d_light_positions}

Interactive 3D control requires the model to interpret 3D position inputs from users, which in \ourmethod\ are tokenized 3D coordinates. While existing works \cite{physically_controllable_relighting_sig_25} typically reconstruct the 3D scene geometry to place lights in actual 3D space, our core design principle is to skip reconstruction and train a model to directly learn scene-light understanding. The key benefit of this design is that the input and interaction are scene-agnostic, though this introduces a challenge in defining the coordinate system. Since users interact on a 2D image canvas, we define 3D coordinates relative to a camera. To implement this, we establish a reference coordinate system consisting of a camera and a 3D sampling volume. All scale-dependent lighting parameters (e.g., position, size, intensity) are defined relative to this volume.

In particular, we define a cube that can be placed in the scene at any scale or orientation. This cube establishes both the light sampling volume and the camera position relative to it. We define a canonical space local to the cube where all lighting parameters are sampled and the camera is fixed. By applying 3D similarity transformations $\in \text{Sim}(3)$ to the cube, we enable the camera to capture any 3D region in the scene. Given two scenes $S_1$ and $S_2$ where $S_2$ is a similarity-transformed version of $S_1$, we can configure the light and camera such that the renderings appear nearly identical.

In the reference space, we set the camera at position $\mathbf{p}_{\text{cam}}$, we sample light at position $\mathbf{p}_{\text{light}}$ with energy $E$, and radius (diffuse level) $d$. Denoting the reference volume origin (center of cube) as $C$, the transformed cube center as $C_t$, and scale factor as $s$, the transformed parameters are computed as follows: the camera position becomes $\mathbf{p}'_{\text{cam}} = C_t + s \cdot (\mathbf{p}_{\text{cam}} - C)$, the light position becomes $\mathbf{p}'_{\text{light}} = C_t + s \cdot (\mathbf{p}_{\text{light}} - C)$, the light intensity scales to $E' = s^2 \cdot E$ to compensate for inverse-square falloff, and the light radius scales to $d' = s \cdot d$ to maintain apparent angular size. This transformation ensures that the rendered appearance remains invariant under similarity transformations of the scene. In practice, we render the scene using the transformed camera and light parameters $(\mathbf{p}'_{\text{cam}}, \mathbf{p}'_{\text{light}}, E', d')$, while tokenizing the lighting attributes using the canonical reference parameters (e.g.,  $\mathbf{p}_{\text{light}}$ and $d$) for model input. For details on rotations, translations, and camera intrinsics, and illustrative diagrams, please see the supplementary.
    
\subsection{Modeling Lighting Edits with Diffusion Model}
\label{subsec:flow_matching_model}

\begin{figure}[!tbp]
    \centering
    \includegraphics[width=0.45\textwidth]{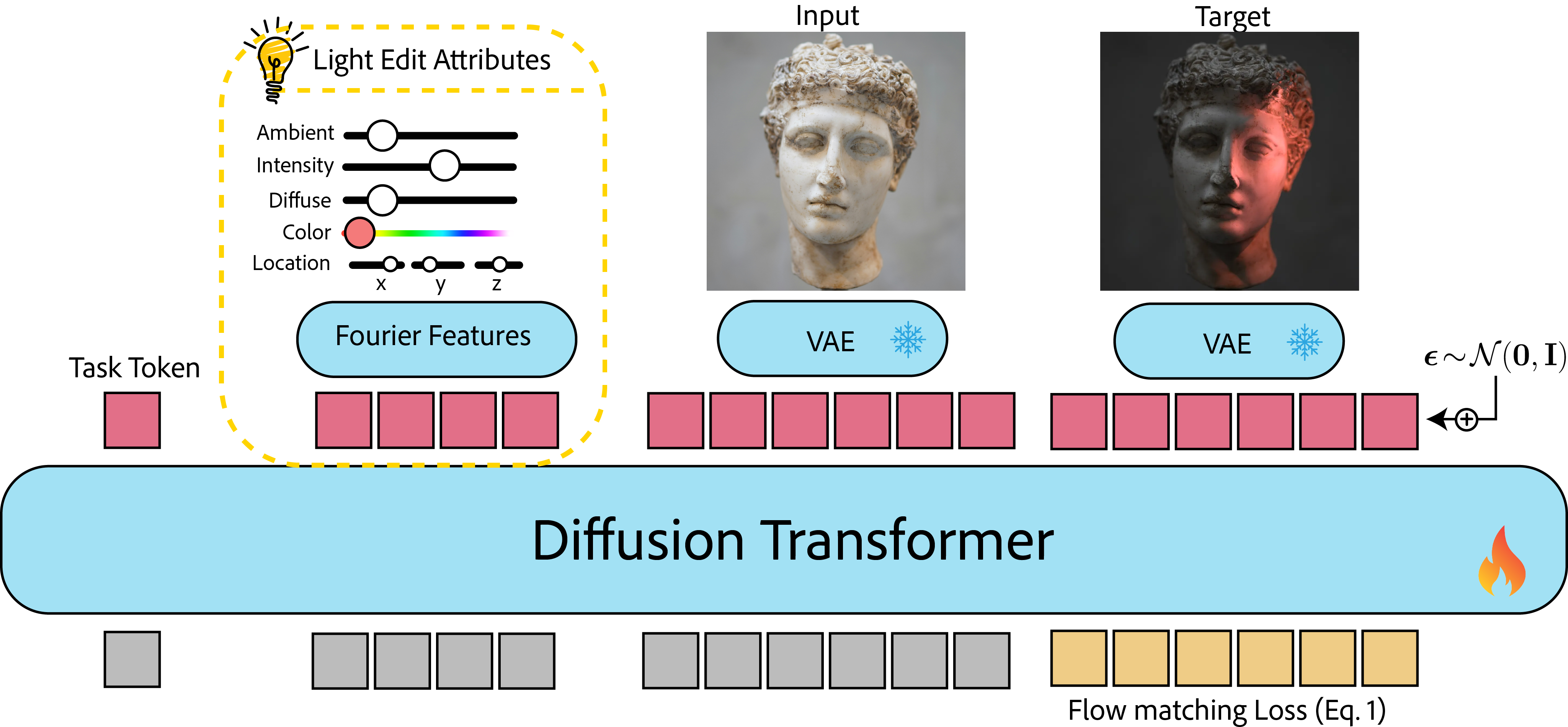}
    \caption{Architecture overview, shown here for the 3D light–position training case. The model takes an input image and light edit attributes; e.g., to add a red light while dimming ambient illumination. For other light attributes, see \cref{subsec:flow_matching_model}.}
    \label{fig:architecture_diagram}
\end{figure}        

We model the conditional distribution $p(I_r | I, \Delta L)$ using a latent diffusion transformer \cite{Peebles_2023_ICCV, rombach2021highresolution} that predicts how an input image changes under a specified lighting edit. 
As depicted in Fig. \ref{fig:architecture_diagram}, both the encoded image tokens and the lighting-edit tokens are concatenated into a single sequence and processed jointly with full self-attention \cite{attention_is_all_you_need}, allowing the transformer to reason about spatial content and lighting control within the same token space. 

\vspace{-2mm}

\paragraph{Lighting Attribute Tokens.}
Scalar lighting attributes in $\Delta L$---such as ambient scaling $a$, intensity $\lambda$, diffuse controls $(d, d_g)$, and the transition flag $t$---are each encoded into tokens using Gaussian Fourier features~\cite{tancik2020fourfeat} with a per-attribute projection matrix $\mathbf{B}\!\sim\!\mathcal{N}(0,\sigma^2)$. Vector attributes, i.e., 3D position $\mathbf{p} = (x, y, z)$ and color $\mathbf{c} = (r, g, b)$, are flattened into one token per component, yielding a uniform token structure across all lighting parameters. The light mask $\mathbf{m}$ is encoded into image latent space via the VAE encoder and flattened into a token sequence. Finally, the input image tokens are concatenated with the $\Delta L$ tokens forming the full conditioning sequence. These are concatenated to the noisy target latent (see \cref{fig:architecture_diagram}). Corresponding image patches share identical positional embeddings~\cite{rope_jsu_2024}.

\vspace{-2mm}

\paragraph{Training objective.}
The model is trained using a flow-matching objective~\cite{lipman2022, lipman2024flow}, which learns a time-dependent velocity field that transports noisy latents toward data latents along straight-line paths. Conditioning is provided by the input image $I$ and its associated lighting edit $\Delta L$. Given a data latent $\mathbf{X} \!\sim\! p(\cdot \mid I, \Delta L)$ and base noise $\boldsymbol{\epsilon} \!\sim\! \mathcal{N}(\mathbf{0}, \mathbf{I})$, we form the interpolant $\mathbf{z}_\tau = (1 - \tau)\,\boldsymbol{\epsilon} + \tau\,\mathbf{X}$, where $\tau \in [0, 1]$. The network $u_\theta(\mathbf{z}_\tau, \tau, I, \Delta L)$ is trained to predict the target velocity $\mathbf{X} - \boldsymbol{\epsilon}$, minimizing:

\begin{equation}
\label{eq:cfm}
\mathcal{L}(\theta)
= \mathbb{E}_{\tau,\,\mathbf{X},\,\boldsymbol{\epsilon}}
\!\left[
\big\|
u_\theta(\mathbf{z}_\tau, \tau, I, \Delta L)
- (\mathbf{X} - \boldsymbol{\epsilon})
\big\|_2^2
\right].
\end{equation}

\begin{table}[!htbp]
\centering
\scriptsize
\setlength{\tabcolsep}{4pt}
\caption{Synthetic data evaluation, comparing against environment-map--based relighting methods Neural Gaffer \cite{jin2024neural_gaffer} and DiffusionRenderer \cite{diffusion_renderer_cvpr_25}. Best results in bold.}
\begin{tabular}{l l ccc}
\toprule
 & \textbf{Method} & \textbf{PSNR} $\uparrow$ & \textbf{SSIM} $\uparrow$ & \textbf{LPIPS} $\downarrow$ \\
\midrule
\textbf{PanoGT}  & DiffusionRenderer \cite{diffusion_renderer_cvpr_25} & 13.5106 & 0.9395 & 0.0394 \\
  & Neural Gaffer \cite{jin2024neural_gaffer}     & 16.7183 & 0.9446 & 0.0412 \\
  & \ourmethod               & \textbf{21.2416} & \textbf{0.9724} & \textbf{0.0228} \\
\midrule
\textbf{PointGT} & DiffusionRenderer \cite{diffusion_renderer_cvpr_25} & 13.5338 & 0.9374 & 0.0402 \\
 & Neural Gaffer \cite{jin2024neural_gaffer}     & 16.7595 & 0.9436 & 0.0415 \\
 & \ourmethod               & \textbf{21.9772} & \textbf{0.9759} & \textbf{0.0210} \\
\hdashline
 & PanoGT             & 31.5661 & 0.9897 & 0.0077 \\
\bottomrule
\end{tabular}
\label{tab:metrics_maskedpsnr_avg}
\end{table}

\begin{figure}[!ht]
    \centering
    \includegraphics[width=0.45\textwidth]{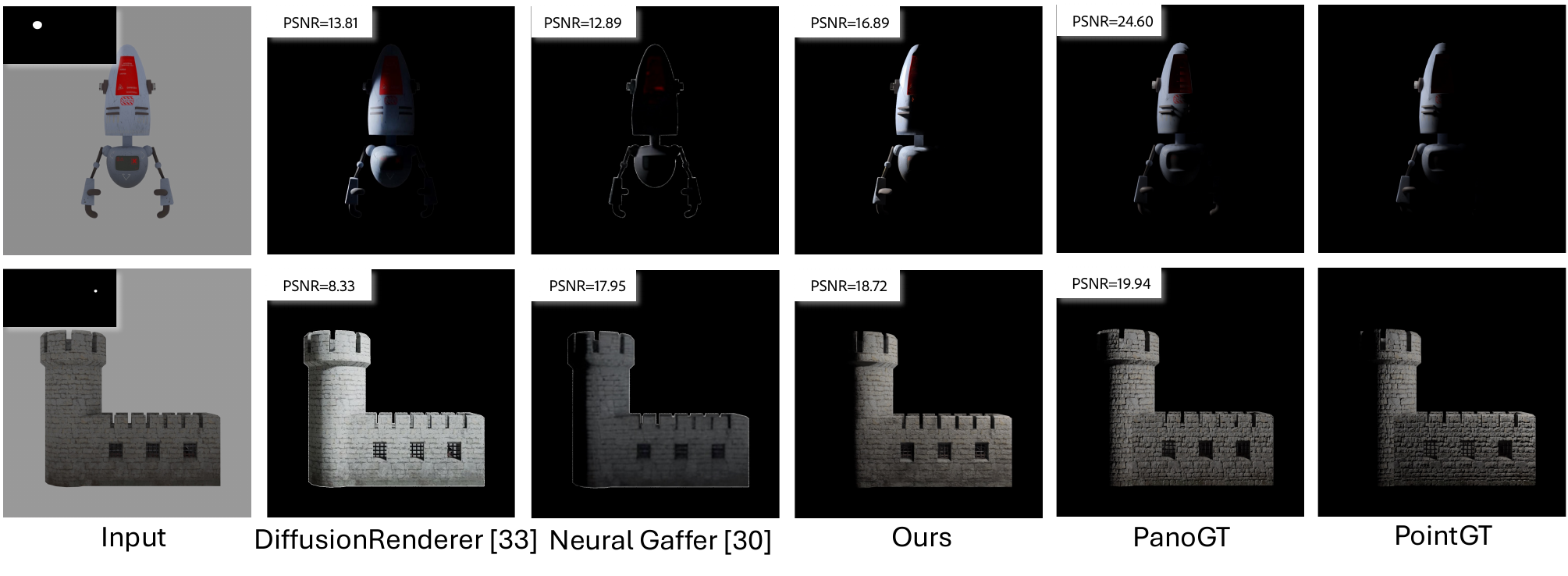}
    \caption{Synthetic data evaluation with Neural Gaffer \cite{jin2024neural_gaffer} and DiffusionRenderer \cite{diffusion_renderer_cvpr_25}. Ground-truth relighting is provided both as point-light renders and as point-lights projected into environment maps, and we report results for both settings.}
    \label{fig:spatial_quantitative}
\end{figure}

\subsection{Implementation Details}

We initialize from a pretrained text-to-video checkpoint and train in bfloat16 using FSDP \cite{fsdp_vldb_2023}. We use AdamW \cite{loshchilov2018decoupled_adamw} (lr $=10^{-5}$, weight decay \(=0.01\), \(\beta\ =(0.9,0.95)\)), a global batch size of \(160\), and \(960\)-px inputs. Training runs for \(15,000\) steps (\(\approx\)2 days) on 16\(\times\)A100-80GB GPUs. Inference uses DDIM sampler \cite{song2021denoising_ddim} with 50 steps. Classifier-free guidance \cite{ho2021classifierfree} is applied only to the light-token conditioning (not input portrait/masks): we form the unconditional branch by dropping \(\Delta L\) and use guidance scale \(w=2\). We set the scale of Gaussian Fourier features \(\sigma = 5\).

\vspace{-2mm}

\section{Experiments}
\label{sec:experiments}

\vspace{-2mm}

~\cref{subsec:quantitative_eval} presents quantitative results: (1) a synthetic dataset for analyzing spatial relighting (2) precision of spatial lighting (3) user study on spatial lighting and (4) evaluation on a real-world captured dataset with controllable light fixtures. ~\cref{subsec:qualitative_comparisons} provides qualitative comparisons with prior methods. ~\cref{subsec:in_the_wild_qualitative} shows in-the-wild results and applications.

\subsection{Quantitative Evaluation}
\label{subsec:quantitative_eval}

\paragraph{Synthetic Evaluation} We evaluate \ourmethod's performance on spatial lighting control on a synthetic benchmark, against two environment-map-based relighting methods:Neural Gaffer \cite{jin2024neural_gaffer} and DiffusionRenderer \cite{diffusion_renderer_cvpr_25}. We randomly sample 200 Objaverse objects from our test set. Each sample has (i) input render under ambient lighting and (ii) two targets where the subject is rendered with either a \emph{point light} (PointGT) or its \emph{environment-map surrogate} (PanoGT), described below. We report metrics PSNR, SSIM, LPIPS \cite{lpips_zhang_cvpr_2018} computed on the masked object regions.

We render point lights as environment maps for environment-map baselines and render objects under these maps to obtain the PanoGT targets, which serve as a fairer ground truth for environment-map methods.
Since point lights induce inverse-square falloff, self-occlusion, and position-dependent penumbrae---unlike the distant, directionally uniform illumination assumed by environment maps---we render scenes at multiple scales. Shrinking the object relative to the light source reduces near-field effects, better matching the environment-map regime. 

\begin{table}[!htbp]
    \centering
    \begin{minipage}[b]{0.52\linewidth}
    \centering
    \scriptsize
    \setlength{\tabcolsep}{1pt}
    \caption{\textit{User Study.}}
    \begin{tabular}{lcc}
    \toprule
     & GenLit \cite{genlit_sig_asia_shrisha_2025} & Careaga \cite{physically_controllable_relighting_sig_25} \\
    \midrule
    Ours preferred & \textbf{77.5}\% & \textbf{89.2}\% \\
    & & \\
    \bottomrule
    \end{tabular}
    \vspace{-3mm}
    \label{tab:userstudy_pref}
    \end{minipage}\hfill
    \begin{minipage}[b]{0.44\linewidth}
    \centering
    \scriptsize
    \setlength{\tabcolsep}{1pt}
    \caption{\textit{Precision Analysis.}}
    \begin{tabular}[t]{lcc}
    \toprule
    \textbf{Method} & $B/A~\uparrow$ & $A~\downarrow$ \\
    \midrule
    Neural Gaffer \cite{jin2024neural_gaffer} & 1.111 & 0.160 \\
    \ourmethod & \textbf{1.877} & \textbf{0.049} \\
    \bottomrule
    \end{tabular}
    \vspace{-3mm}
    \label{tab:precision_analysis}
    \end{minipage}
\end{table}
\begin{figure}[!htbp]
\centering
\tiny
\includegraphics[width=\linewidth]{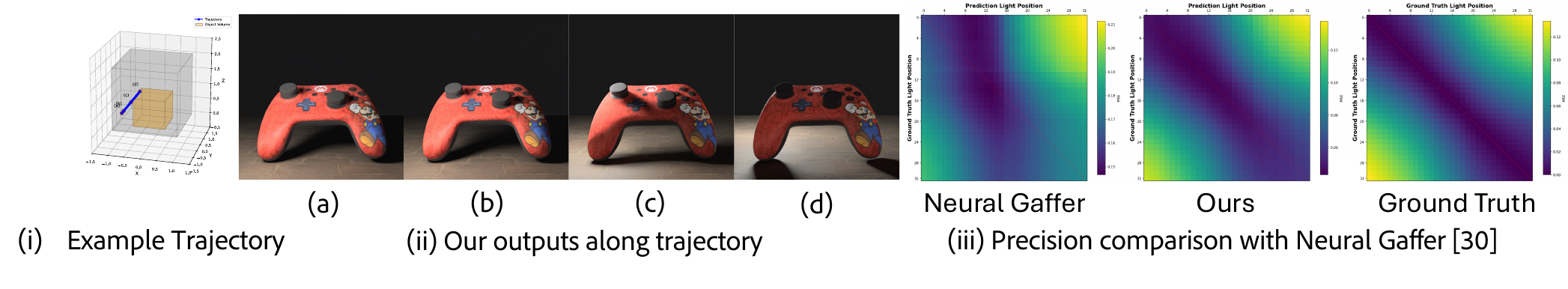}
\caption{\textit{Example trajectory and confusion maps} measure precision as sensitivity ($B/A\uparrow$) and accuracy ($A\downarrow$). Our method is closer to ground truth than Neural Gaffer \cite{jin2024neural_gaffer}.}
\label{fig:precision_q}
\end{figure}

\begin{table}[!htbp]
\centering
\scriptsize
\setlength{\tabcolsep}{6pt}
\caption{Quantitative results on VisibleFixture-60, comparing against ScribbleLight \cite{scribble_light_choi_2025_cvpr}. Best results in bold.}
\begin{tabular}{lccc}
\toprule
\textbf{Method} & \textbf{PSNR}$\uparrow$ & \textbf{SSIM}$\uparrow$ & \textbf{LPIPS}$\downarrow$ \\
\midrule
ScribbleLight \cite{scribble_light_choi_2025_cvpr} & 14.6356 & 0.5186 & 0.6114 \\
\ourmethod & \textbf{20.0772} & \textbf{0.8476} & \textbf{0.2784} \\
\bottomrule
\end{tabular}
\label{tab:unified_results_visible_only}
\end{table}

\begin{figure}[!htbp]
    \centering
    \includegraphics[width=0.45\textwidth]{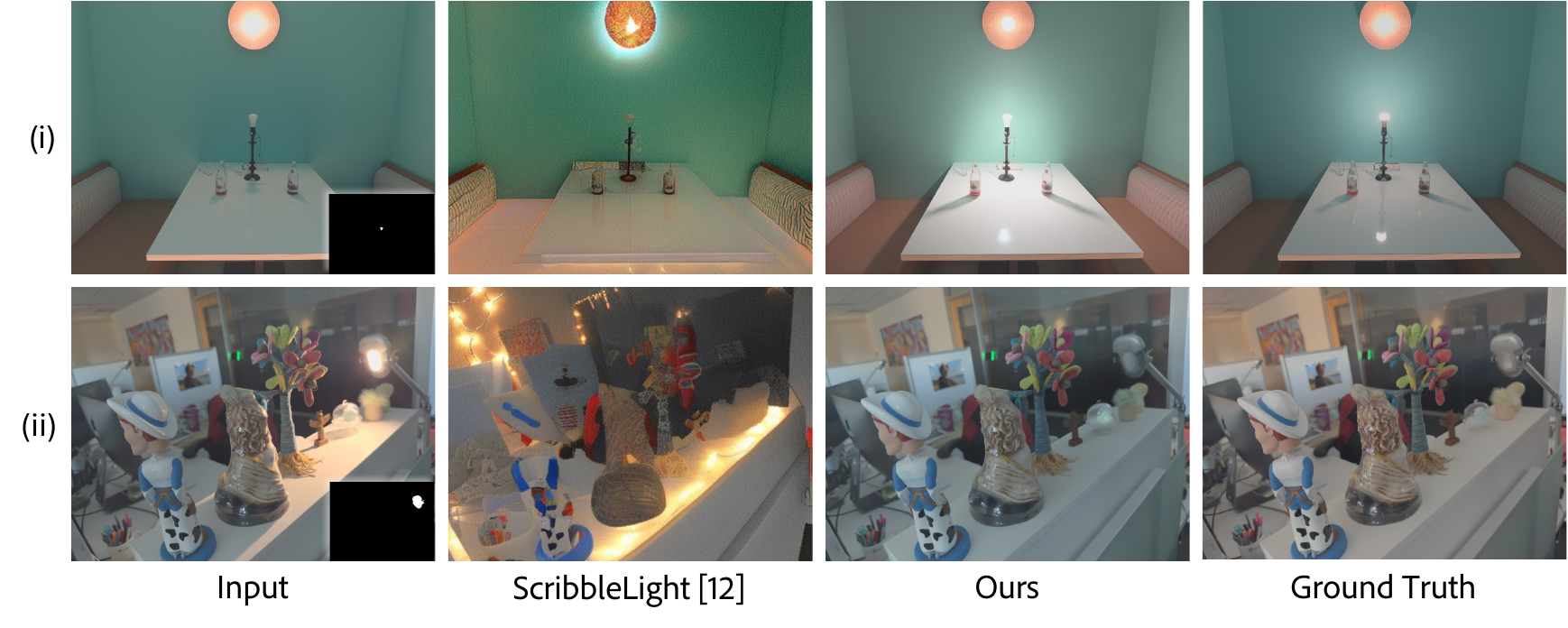}
    \caption{VisibleFixture-60. We show two cases: (i) turning a light on produces reflections and shadows that closely match the ground truth; (ii) turning the light off removes all cast shadows.}
    \label{fig:real_world_capture}
\end{figure}
\begin{figure*}[!htbp]
    \centering


    \begin{subfigure}{\textwidth}
        \centering
        \includegraphics[width=\linewidth]{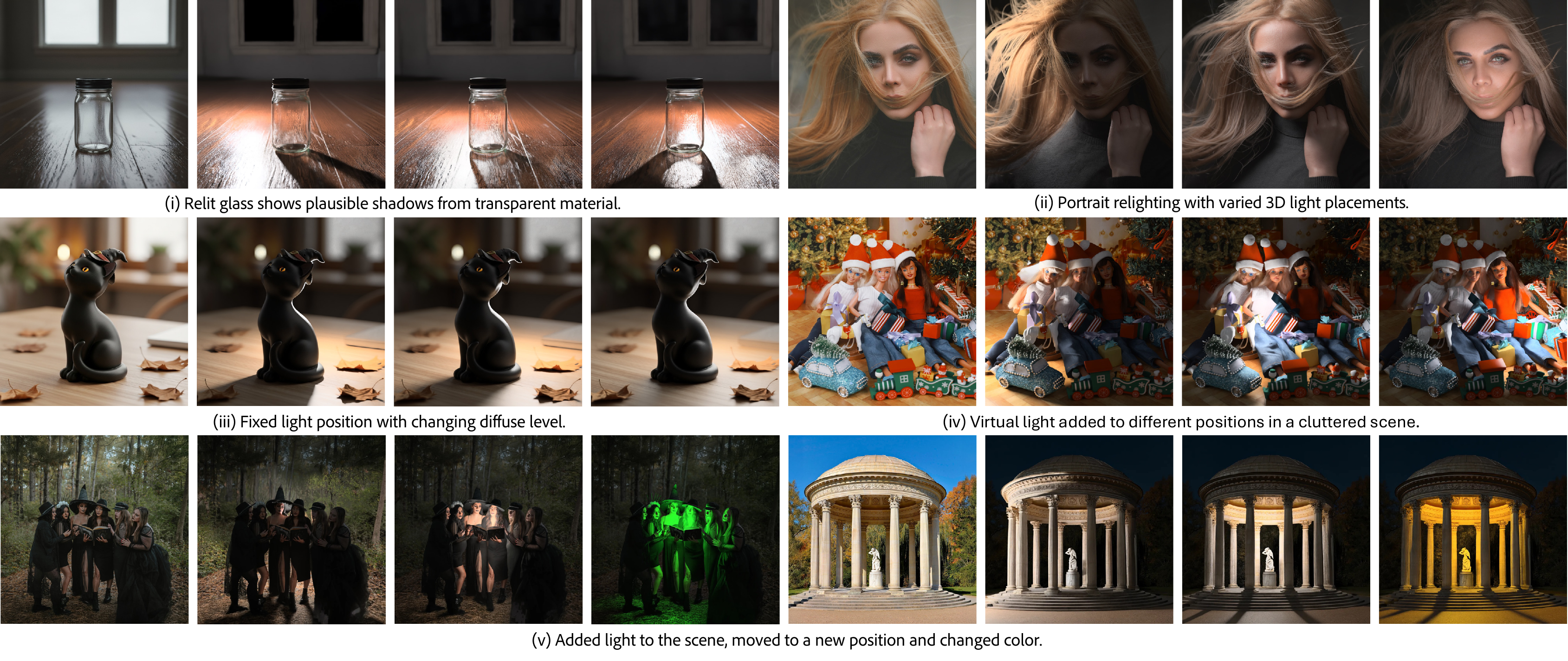}
        \caption{Spatial control in in-the-wild testing cases. For each example we show input (left) and three relit results (right).}
        \label{fig:spatial_results}
    \end{subfigure}

    \begin{subfigure}{\textwidth}
        \centering
        \includegraphics[width=\linewidth]{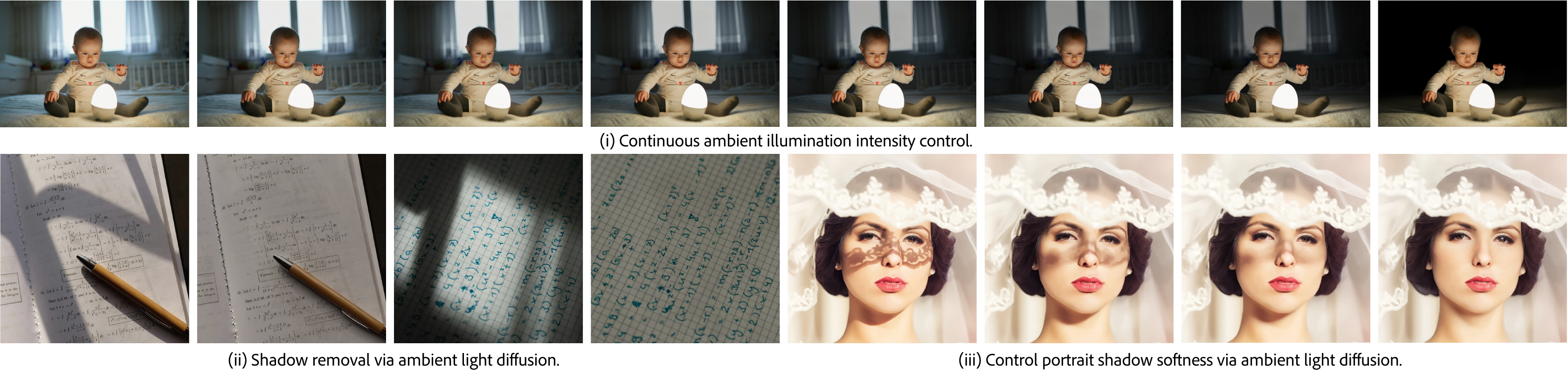}
        \caption{Each example shows the input on the left and relit outputs on the right, demonstrating environment-lighting control through ambient dimming, document de-shadowing, and progressive increases in global diffuse level in portraits.}
        \label{fig:diffuse_results}
    \end{subfigure}

    \begin{subfigure}{\textwidth}
        \centering
        \includegraphics[width=\linewidth]{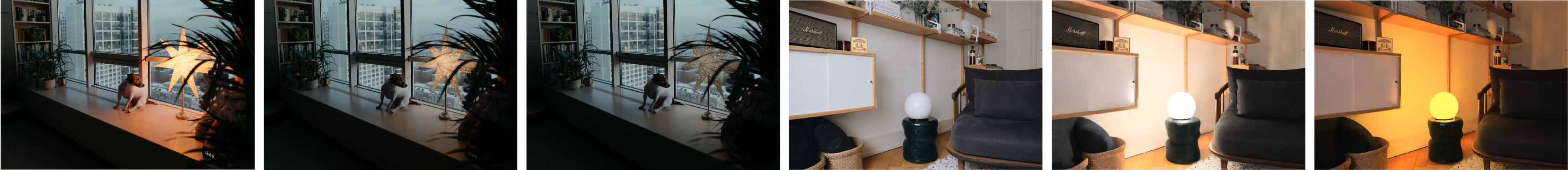}
        \caption{Examples showing fine-grained manipulation of visible light intensity and color enabling precise per-light adjustments.}
        \label{fig:intensity_results}
    \end{subfigure}

    
    \caption{In-the-wild examples showcasing the method’s precise lighting control. Best viewed on a screen—zoom in for details.}
    \label{fig:ours_qualitative}
    
    \vspace{-4mm}
    
\end{figure*}
\begin{figure}[!ht]
    \centering
    \includegraphics[width=0.47\textwidth]{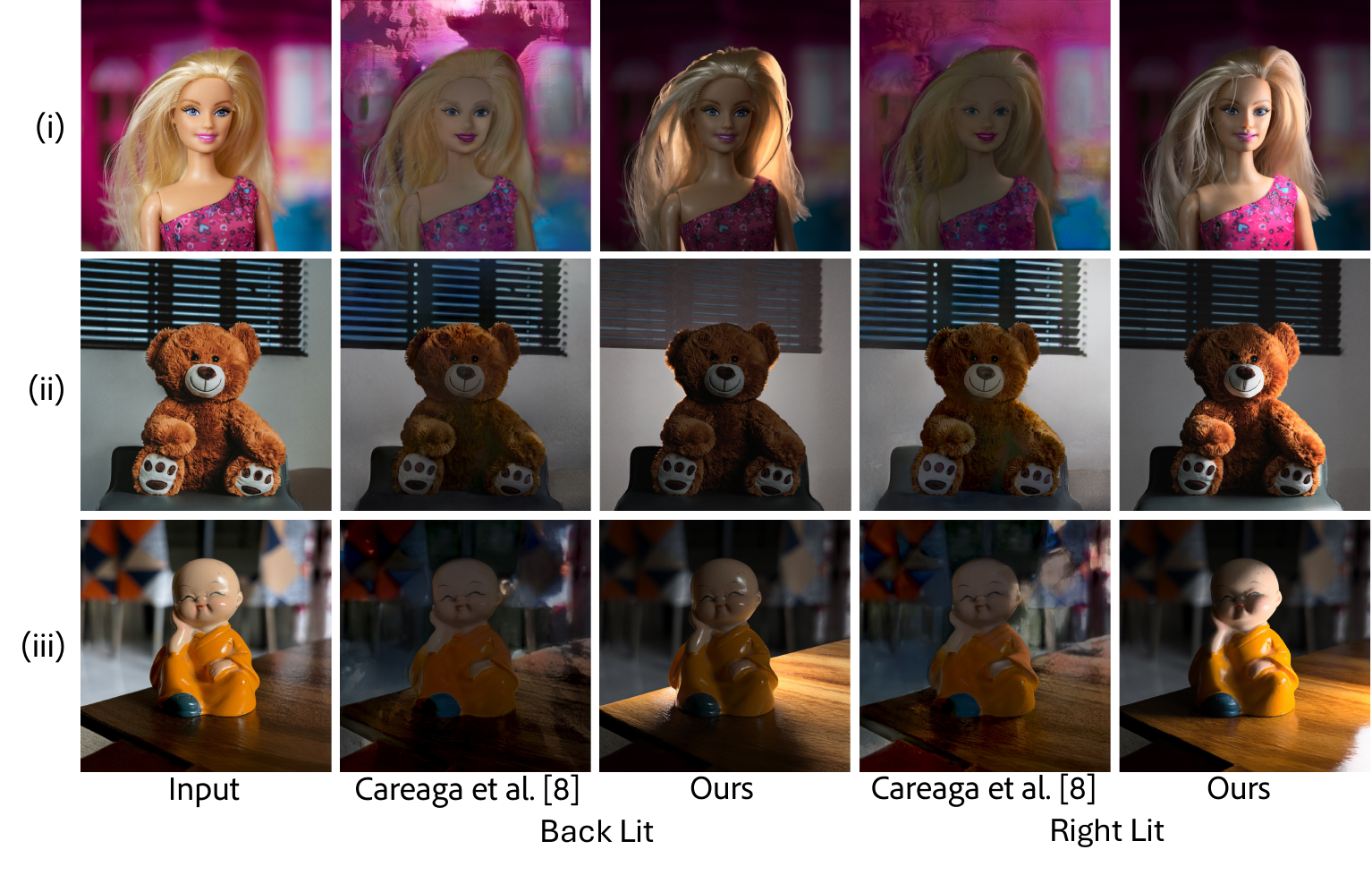}
    \caption{Relighting comparison with Careaga et al. \cite{physically_controllable_relighting_sig_25}. (i) Backlit Barbie example where the light source produces a rim; (ii) a teddy-bear scene illustrating realistic relighting of fine fur; and (iii) a porcelain statue showing specular highlights.}
    \label{fig:careaga_comparison}
\end{figure}

\begin{figure}[!ht]
    \centering
    \includegraphics[width=0.47\textwidth]{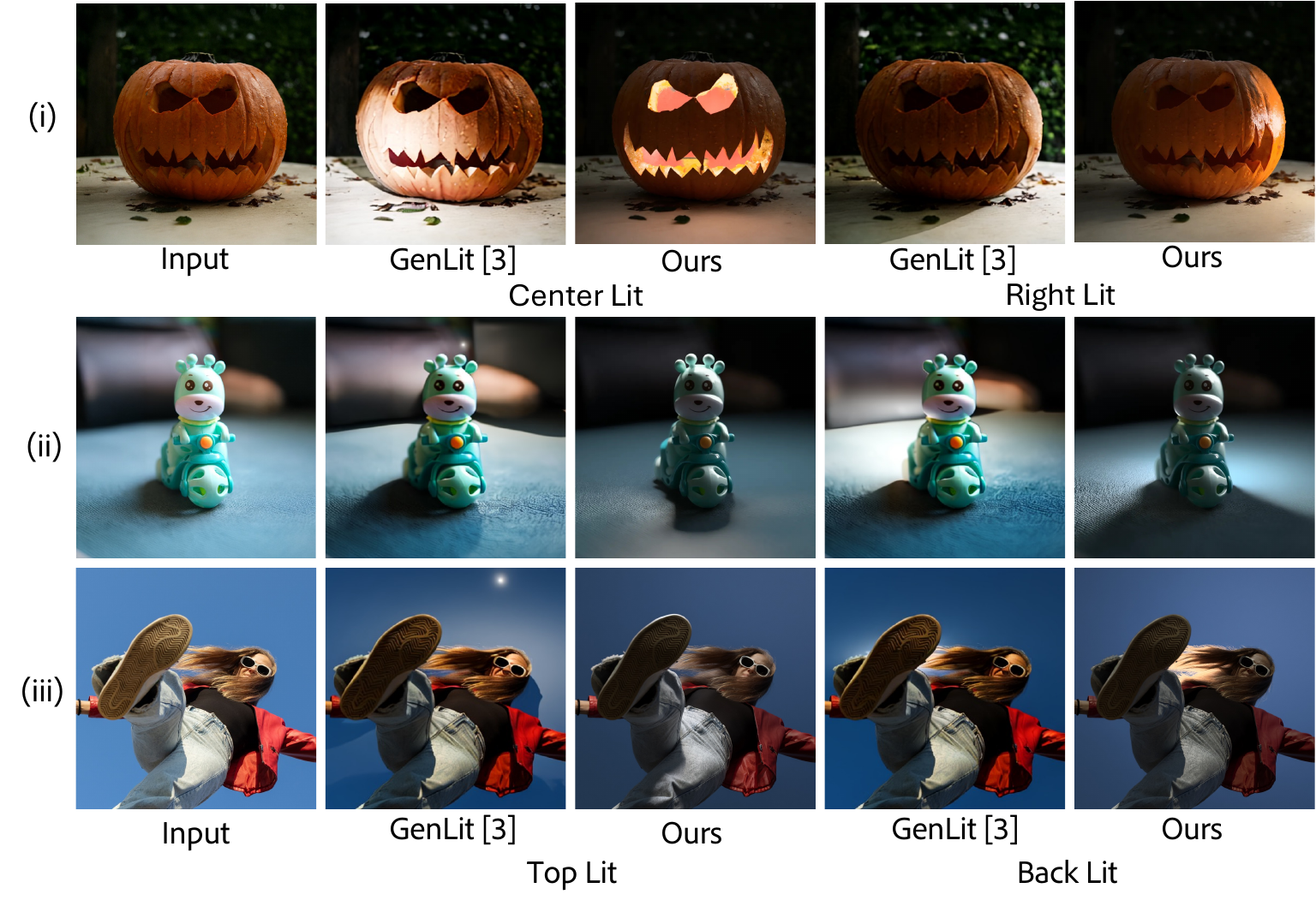}
        \caption{Comparison with GenLit \cite{genlit_sig_asia_shrisha_2025}: (i) Centered pumpkin light reproduced only by ours; (ii) Frontal view where both place the top light correctly; (iii) Low-angle view where GenLit’s top light drifts, while ours remains consistent.}
    \label{fig:comparison_with_genlit}
\end{figure}


We report results averaged over all scales. \ourmethod\ outperforms baselines on PointGT and PanoGT targets (see \cref{tab:metrics_maskedpsnr_avg}, \cref{fig:spatial_quantitative}), demonstrating strong spatial lighting control.

\vspace{-2mm}

\paragraph{Precision Analysis}

To evaluate precision of \ourmethod\ to light positions, we construct a benchmark with 50 Objaverse test objects. For each object, we evaluate six linear point-light trajectories (32 steps) along three axes (two trajectories/axis). Given an ambient input, we predict relit images for each light position, and compare against ground-truth renders using foreground-masked MSE.

To measure how well the model distinguishes nearby light positions, we create a confusion matrix $M \in \mathbb{R}^{32 \times 32}$, where $M_{ij}$ denotes the error between the prediction conditioned on light position $j$ and the ground truth (GT) rendered at light position $i$. $M$ captures both \emph{accuracy} (low diagonal error) and \emph{sensitivity} (off-diagonal error increasing with trajectory distance). In \cref{tab:precision_analysis}, we report $A$ (lower is better) and $B/A$ (higher is better), where $A, B$ are:

\begin{equation}
    A = \frac{1}{32}\sum_{i=1}^{32} M_{ii} \ ,\quad B = \frac{\sum_{i\neq j}|i-j|M_{ij}}{\sum_{i\neq j}|i-j|}
\end{equation}

In \cref{fig:precision_q}, we show results along an example trajectory and the confusion maps. We include a GT vs.\ GT reference. Our method achieves lower $A$, higher $B/A$, and confusion maps closer to GT than Neural Gaffer, indicating more precise spatial light control.

\vspace{-2mm}

\paragraph{User Study}

We evaluate spatial lighting via a user study on in-the-wild images where participants select results that better match target lighting reference with higher image quality. \ourmethod\ is preferred over both GenLit and Careaga et al. (\cref{tab:userstudy_pref}). Details are in the supplementary.

\vspace{-2mm}

\paragraph{Visible-fixture pairs (in-the-wild).}

We evaluate on 60 real-world captured on/off image pairs where individual light fixtures can be toggled, with binary masks indicating each fixture. We construct a test set called VisibleFixture-60 and report PSNR, SSIM, LPIPS, comparing with ScribbleLight~\cite{scribble_light_choi_2025_cvpr}. Each fixture is tested in both directions (off$\rightarrow$on and on$\rightarrow$off), by swapping input and target images and conditioning both models accordingly (see~\cref{tab:unified_results_visible_only,fig:real_world_capture}).

Our method produces visually consistent and physically plausible relighting effects that closely match ground truth. In \cref{fig:real_world_capture}(i), the light reflects naturally off the table and casts shadows from the bottles. In \cref{fig:real_world_capture}(ii), turning the light off correctly removes all corresponding shadows, demonstrating accurate light-geometry interaction.

\subsection{Qualitative Comparisons}
\label{subsec:qualitative_comparisons}

We compare with recent relighting methods: GenLit \cite{genlit_sig_asia_shrisha_2025}, Careaga et al. \cite{physically_controllable_relighting_sig_25}, and LightLab \cite{light_lab_sig_25}. Since all are closed-source, results for GenLit and Careaga et al. were produced by authors on an in-the-wild test set we created, while LightLab results are from their publication.

GenLit and Careaga et al. are among the few methods supporting spatially controllable relighting. GenLit conditions a diffusion model on global light coordinates while sampling random camera viewpoints during training, coupling camera orientation with spatial light interpretation. Our scene-agnostic design avoids this. In \cref{fig:comparison_with_genlit}, we show three examples: (i) for a centered light on a pumpkin, our method reproduces the lighting while GenLit does not; (ii) ours yields a stronger backlit effect; and (iii) for a low-angle view with the same top-light specification, GenLit’s placement drifts sideways while ours stays consistent, indicating that GenLit’s spatial behavior depends on camera configuration and scene content, causing inconsistent interaction. We provide more stable and predictable spatial control.

Careaga et al. adopt a two-stage pipeline that reconstructs a 2.5D scene representation before neural rendering. This approach often falters in scenes with heavy occlusion or materials exhibiting semi-transparency or view-dependent effects. As shown in \cref{fig:careaga_comparison}, our method produces convincing backlit hair with natural translucency and accurately captures specular highlights on the porcelain figurine, whereas their method struggles with these phenomena.

Our method offers similar results, compared to LightLab, for turning visible light-fixtures on/off. We present this comparison in the supplementary.  

\subsection{In-the-wild Results}
\label{subsec:in_the_wild_qualitative}

\vspace{-1mm}

\paragraph{Add 3D Light in 2D Image.}

We demonstrate a wide range of spatial lighting effects across diverse scenarios in \cref{fig:spatial_results}. Our method handles challenging materials, producing plausible shadows when relighting transparent glass (\cref{fig:spatial_results}(i)) and high-fidelity relighting of hair (\cref{fig:spatial_results}(ii)). It works well with both simple objects (\cref{fig:spatial_results}(iii)) and cluttered scenes (\cref{fig:spatial_results}(iv)). The model's understanding of complex scene geometry enables nuanced light placements, such as positioning a light behind a book to illuminate the portraits of a coven or inside a temple dome (\cref{fig:spatial_results}(v)).

\vspace{-4mm}

\paragraph{Ambient lighting controls.}
We show diverse ambient light controls in \cref{fig:diffuse_results}. In \cref{fig:diffuse_results}(i), progressively turning down the ambient lighting continuously lowers the intensity of background lighting from the window, focusing the scene on the main character. Diffuse ambient lighting enables us to remove shadows on documents cleanly (\cref{fig:diffuse_results}(ii)) or gradually soften shadows on a portrait (\cref{fig:diffuse_results}(iii)).

\vspace{-4mm}

\paragraph{Visible in-scene light intensity controls.}
Our tokenized intensity and color parameters can be combined with masks for in-scene lighting adjustment, enabling realistic lamp on/off effects and color control (\cref{fig:intensity_results}).
\section{Conclusion}
\label{sec:conclusion}

We presented \ourmethod, a diffusion-based framework for image relighting that achieves precise, continuous control over multiple illumination attributes. By encoding lighting parameters such as intensity, color, diffuse level and 3D position into a compact, token-based representation, our method directly maps an input image and lighting edit to a relit output without explicit inverse rendering. Trained on a large-scale synthetic dataset with additional real captures, \ourmethod\ attains state-of-the-art performance across diverse relighting tasks and demonstrates a robust learned understanding of light–scene interaction. We discuss limitations and future work in the supplementary.



\section*{Acknowledgements}

We thank Weijie Lyu for insightful discussions and the participants of our user study. We also thank Shrisha Bharadwaj, Chris Careaga, and Nadav Magar for providing comparisons with their respective methods. Finally, we thank Kalyan Sunkavalli and Nathan Carr for their support.

{
    \small
    \bibliographystyle{ieeenat_fullname}
    \bibliography{main}
}

\appendix

\clearpage
\setcounter{page}{1}

\maketitlesupplementary

\begin{figure*}[!htbp]
    \centering
    \includegraphics[width=\linewidth]{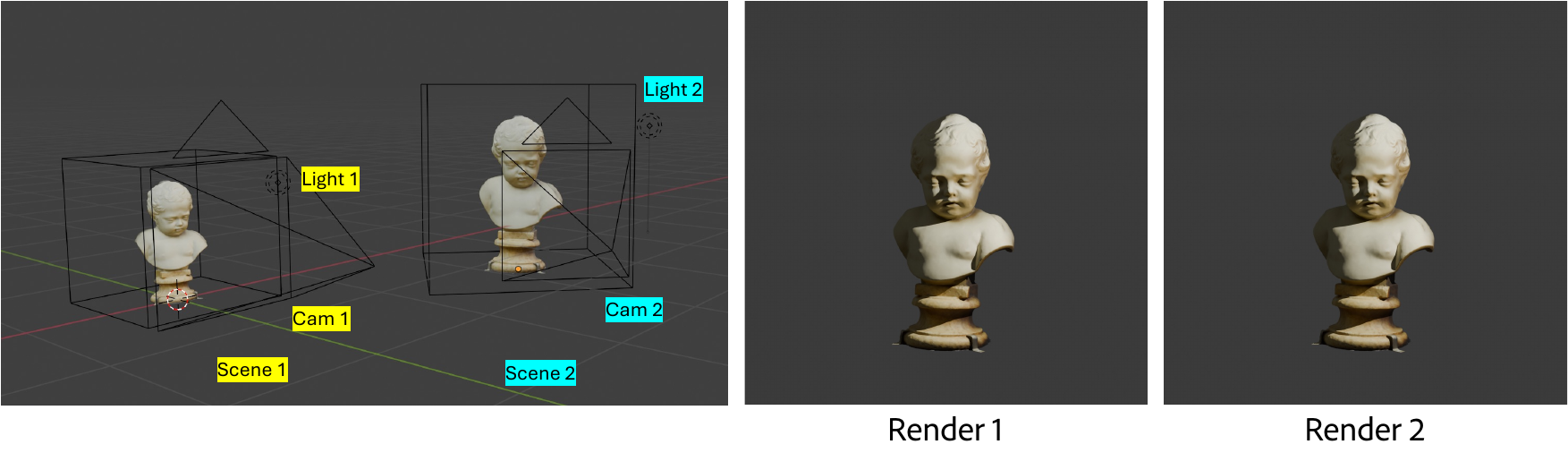}
    \caption{An illustration of the scene-agnostic camera and lighting parameterization, implemented via a transformation from a canonical reference space. Scene 2 is a similarity-transformed version of Scene 1. We map Cam 1 $\rightarrow$ Cam 2 and Light 1 $\rightarrow$ Light 2 using the scene transformation defined in our paper. Rendering Scene 1 with Cam 1 produces Render 1, and rendering Scene 2 with Cam 2 produces Render 2, which is visually indistinguishable from Render 1 (only one scene is rendered at a time). This construction allows cameras and lights to be defined in a canonical reference space while scenes are freely placed at different locations, scales, and orientations, ensuring a consistent relationship between lighting parameters and their visual effects in the image space.}
    \label{fig:tsr_illustration}
\end{figure*}

In the supplementary, we provide additional implementation details, analysis, qualitative results, and comparisons with baseline methods. Please refer to the supplementary video for a visual overview and further qualitative results.

\section{Experimental Setting}
\label{sec:experimental_setting}

\subsection{Synthetic Evaluation}
\label{subsec:synthetic_evaluation}

Here we provide additional details on synthetic evaluation. This complements our description in the main paper in \cref{subsec:quantitative_eval}. 

For the synthetic benchmark, we compare against two ground-truth targets: PointGT, rendered with the original point light, and PanoGT, rendered using an environment-map representation of that light, for fairness in our comparison with environment-map--based baselines. Here we describe how PanoGT is constructed.

To create the environment map, we replace the point light with an emissive sphere of radius $r$ centered at the original light position. The sphere is assigned a uniform emission $L$ such that its total emitted energy matches the point light \cite{pharr2023physically}: 

\begin{equation}
    L = \frac{E}{4\pi^2 r^2}.
\end{equation}

A panoramic camera is then placed at the center of the object to render the environment map, which serves as the lighting input for environment-map baselines. Rendering the object under this map produces the PanoGT target. The panoramic camera captures the incident radiance field $L(\mathbf{x}, \boldsymbol{\omega})$ at location $\mathbf{x} \in \mathbb{R}^3$, where $\boldsymbol{\omega}$ denotes directions on the unit sphere. This provides an approximation to rendering with a 3D point light, as it assumes $L(\mathbf{x}, \boldsymbol{\omega})$ is constant across all surface points in the scene. When the scene is lit by a distant point light, $L(\mathbf{x}, \boldsymbol{\omega})$ varies slowly as $\mathbf{x}$ changes, making this approximation valid. However, the approximation degrades as the point light becomes more local to the scene, where $L(\mathbf{x}, \boldsymbol{\omega})$ exhibits stronger spatial variation. This represents the limitation of using environment map to model spatially varying lighting, and consequently why PanoGT and PointGT exhibit small but non-zero differences, as seen in \cref{tab:metrics_maskedpsnr_avg}, last row.

\subsection{VisibleFixture-60 Capture Protocol}
\label{subsec:visible_fixture_cap_proto}

We capture a dataset of indoor office scenes to evaluate our ability to manipulate lighting from visible light fixtures. We capture a scene with controlled illumination changes by turning the visible light fixture on/off. All captures are performed in office spaces using an iPhone mounted on a tripod, with a fixed camera pose across all shots. We use iPhone's ProRAW mode with auto-exposure enabled. Since the camera remains fixed, no image registration or alignment is required.

The captured ProRAW images are 4K/8K Digital Negative (DNG) images and are converted to EXR format subsequently tone-mapped using the Reinhard operator \cite{reinhard_tone_mapping}. For each scene, we manually segment visible light fixtures to identify controllable sources.

\section{Additional Implementation Details}
\label{sec:implementation_extra}

\paragraph{Paired Supervision Synthesis}

Here we provide additional details on synthesizing training pairs, complementing the description in the main paper in \cref{subsec:synthetic_data}. 

To support intensity, color, spatial, and diffuse controls, we generate supervision on-the-fly during data loading by combining an ambient render $I$, an on-light render $O$, and a tone-mapping operator $\mathbf{T}(\cdot)$.

\paragraph{Visible-Light fixtures} For visible fixtures, we construct the ambient image by adding the environment-map render with contributions from up to five \emph{non-selected} fixtures. We then choose one fixture to act as the controllable source, use its corresponding render as $O$, and combine them as

\begin{equation}
I_r = \mathbf{T}\!\big(a\,I + \lambda\,\mathbf{c}\,O\big),
\end{equation}

where $a$ is the ambient scale, $\lambda$ the intensity (both in $[0,1]$), and $\mathbf{c}$ the sampled color. This setup ensures that only the selected fixture contributes the controllable light, while all remaining illumination is interpreted as ambient, cleanly separating the masked light’s effect from the surrounding lighting.

\paragraph{Spatial lighting.}

For spatial lighting, we follow a parallel construction. The ambient image $I$ is given by the environment-map render, while the controllable source is a point light with a sampled 3D position. The render $O$ corresponds to this single point light, and we again form the relit image as

\begin{equation}
    I_r = \mathbf{T}\!\big(a\,I + \lambda\,\mathbf{c}\,O\big).
\end{equation}

This formulation provides supervision for precise spatial control: the model learns how a point light at a known 3D location influences the scene.

\paragraph{Global diffuse-level control}
For diffuse-spread supervision, we fix the light position and render multiple spread levels. Let $A$ denote the ambient render, $O_1$ and $O_2$ the on-light renders with different spread parameters, and $I$ and $I_r$ the corresponding tone-mapped input and target images. Training pairs hold intensity, color, and location constant while varying only the spread parameter:
\begin{equation}
\begin{split}
I   &= \mathbf{T}\!\big(a\mathbf{c}_1\,A+\lambda\,\mathbf{c}_2\,O_1\big),\\
I_r &= \mathbf{T}\!\big(a\mathbf{c}_1\,A+\lambda\,\mathbf{c}_2\,O_2\big).
\end{split}
\end{equation}
The difference in spread between $O_1$ and $O_2$, $d_g$ conditions the model. Note that in order to ensure that perceived differences in shadow softness only stem from $O_1$ and $O_2$ we use constant ambient lighting for $A$, avoiding environment map renders as they can introduce strong directional shadows (e.g., from sunlight) that would appear in both $I$ and $I_r$, breaking the intended spread-only supervision, since such fixed shadows would confound the effect of changing only the diffuse-spread parameter.

In all three cases, the network predicts $I_r$, given $I$ and lighting-edit attributes. 

\paragraph{Scene-Agnostic Camera and Lighting}
Here, we provide additional details for our implementation of scene-agnostic camera and lighting that complement the description in the main paper in \cref{subsec:representing_3d_light_positions}.

We define a canonical reference space containing a camera at position $\mathbf{p}_{\text{cam}}$ and a light sampling volume centered at $C$. Light parameters, 3D position $\mathbf{p}_{\text{light}}$, energy $E$, and radius $d$, are drawn from this space. The cube extends both in front of and behind the image plane, enabling the sampling of lights that fall partially or fully outside the visible frame. All renders use a fixed field of view of $39.6^\circ$ across our synthetic scenes. 

During testing, the field of view (FOV) of the input images is unknown. We find that the effectiveness of 3D position controls is not affected, due to the use of normalized coordinates in the reference space.

To expose the model to viewpoint variability while preserving the meaning of the canonical coordinates, we apply 3D similarity transformations to both the lights and the camera during data generation.

Given a target cube center $C_t$ and scale factor $s$, we first apply a translation that moves both $\mathbf{p}_{\text{cam}}$ and $\mathbf{p}_{\text{light}}$ from the canonical center $C$ to $C_t$:
\begin{equation}
\mathbf{p}_{t} = \mathbf{p} + (C_t - C).
\end{equation}

We then apply a uniform scaling about the new center:
\begin{equation}
\mathbf{p}_{ts} = C_t + s \, (\mathbf{p}_{t} - C_t).
\label{eq:position_scaling}
\end{equation}

Scaling also adjusts appearance-dependent parameters. The light energy becomes
\begin{equation}
E_{s} = s^2 E,
\label{eq:energy_scaling}
\end{equation}
which preserves perceived brightness by compensating for inverse-square falloff, and the radius becomes
\begin{equation}
d_{s} = s d,
\label{eq:radius_scaling}
\end{equation}
which maintains the angular extent of the light as viewed from the camera. Together, these adjustments ensure that the apparent lighting behavior is invariant under uniform rescaling of the scene.

We optionally apply a rotation $R \in SO(3)$ about $C_t$:
\begin{equation}
\mathbf{p}_{tsr} = C_t + R(\mathbf{p}_{ts} - C_t).
\end{equation}
This allows the model to observe the same canonical configuration under diverse orientations while preserving the geometry of camera–light relationships.

In practice, rendering uses the fully transformed parameters $(\mathbf{p}_{tsr,\text{cam}}, \mathbf{p}_{tsr,\text{light}}, E_s, d_s)$, while the model is conditioned only on the \emph{canonical} (pre-transform) parameters $(\mathbf{p}_{\text{light}}, d)$, which are the values provided by users at inference. This separation ensures that tokenized 3D coordinates correspond to a consistent spatial meaning irrespective of the underlying scene content or the particular similarity transform applied during training. 

We provide a visualization in \cref{fig:tsr_illustration} to illustrate how the scene transformation operates. The figure shows two scenes, each containing a camera and a point light, related by a similarity transform. We render both scenes after transforming the light attributes and camera locations under our formulation. The resulting images are visually indistinguishable, confirming that our parameterization preserves lighting behavior under arbitrary scene placement. This enables all lighting parameters to be defined once in a canonical space and ensures that lighting parameters produce consistent visual effects in image space as scenes are placed freely at different locations, scales and orientations.

\paragraph{Additional Training Details}

We use a 2B-parameter diffusion transformer. Each 960\,px image is encoded into a latent of shape \([12,120,120]\), patchified \(2\times2\) to \([48,60,60]\), projected to 4096-dimensional tokens, and flattened into a 3600-token sequence. Training runs on two nodes (each with 8\(\times\)A100\,80GB) with a total batch size of 160, split as 64:48:48 across visible-light, spatial, and diffuse cases. To enable classifier-free guidance at inference, we drop lighting edit $\Delta L$ tokens 10\% of the times, replacing them with a tensor of matching sequence length and dimensionality with values -1. 

\section{Additional User Study Details}

\begin{figure}[!htbp]
    \centering
    \includegraphics[width=0.45\textwidth]{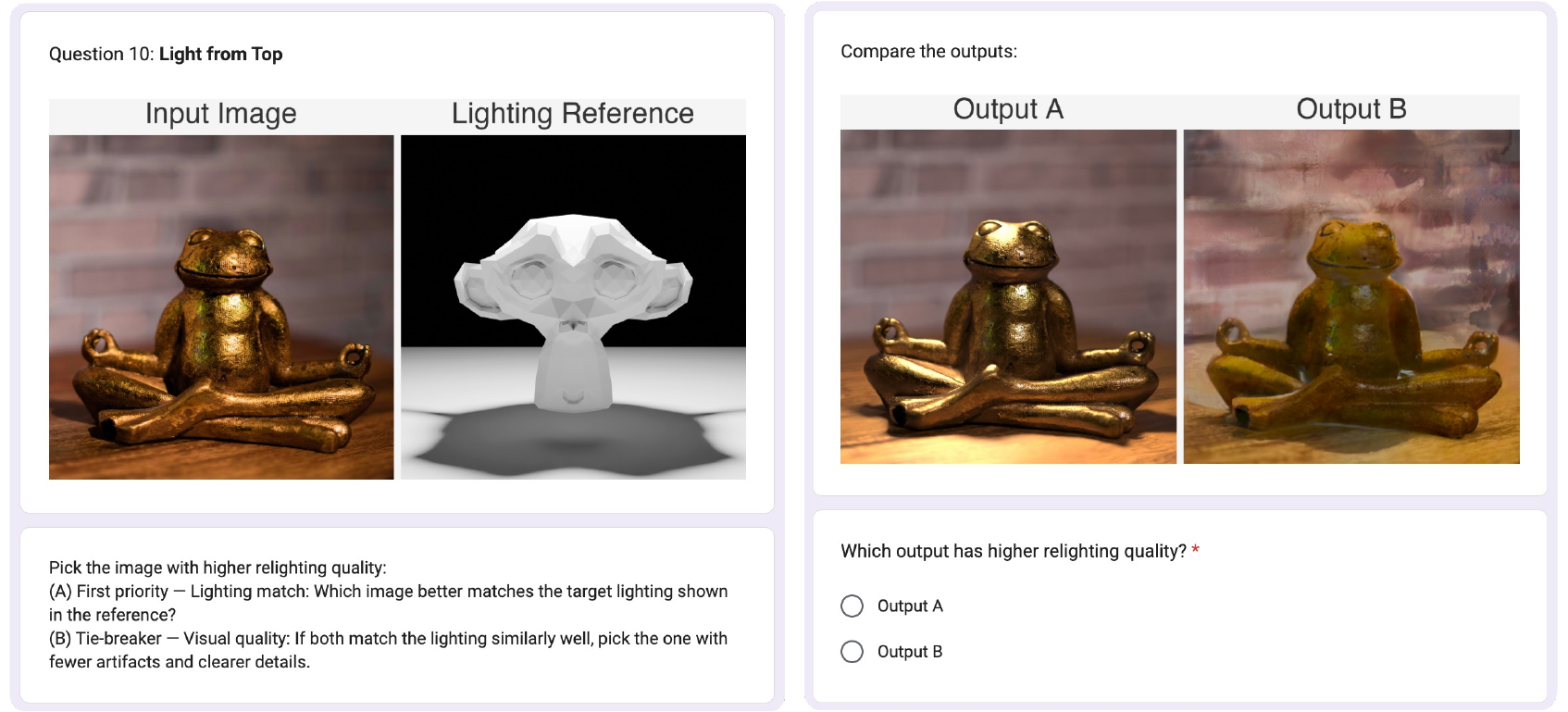}
    \caption{\textit{User Study Questionnaire.}}
    \label{fig:user_study}
\end{figure}

Here, we provide additional details for the user study summarized in \cref{tab:userstudy_pref} of the main paper. In the absence of ground-truth lighting, quantitative evaluation on in-the-wild images is challenging. We therefore conduct a user study comparing \ourmethod\ against the closest spatial-control baselines, GenLit \cite{genlit_sig_asia_shrisha_2025} and Careaga et al. \cite{physically_controllable_relighting_sig_25}. As shown in \cref{fig:user_study}, each trial presents an input image, a target-lighting reference, and two relit outputs. Users select the output that better matches the target lighting; if both match the target similarly well, they break ties by choosing the one with fewer artifacts. In the example shown in \cref{fig:user_study}, the input is a metallic frog figurine, and the target-lighting reference is a Suzanne render lit from above. This reference serves as a visual guide for the lighting cues users should attend to, such as shadow direction and the placement of bright regions. In this example, Output A better matches the intended top-down lighting than Output B and exhibits fewer artifacts. Using 20 in-the-wild images and 5 target lights, we collect votes from 18 users for each baseline comparison. We apply Reinhard auto-exposure \cite{reinhard_tone_mapping} to all outputs to reduce perceived brightness as a confounding factor.

\section{Effect of Inference steps on Quality/Speed}

\begin{figure*}
    \centering
    \includegraphics[width=\linewidth]{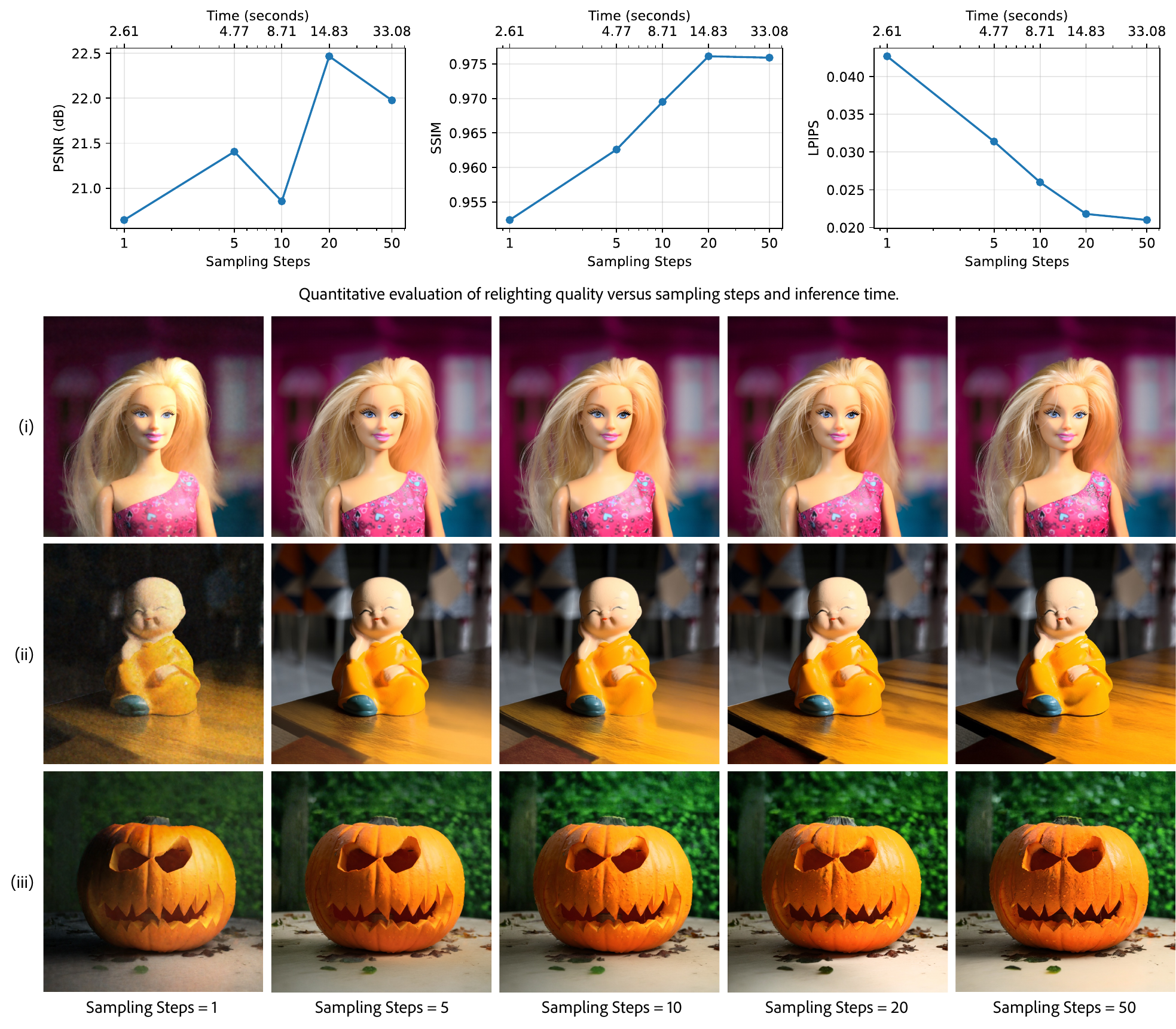}
    \caption{\textit{Effect of sampling steps on relighting quality.} Quantitative evaluation of PSNR, SSIM, LPIPS versus sampling steps and inference time (top). In-the-wild qualitative results 
  at varying step counts for inputs from \cref{fig:careaga_comparison} rows (i), (iii) and \cref{fig:comparison_with_genlit} (i). (bottom).}
    \label{fig:inference_scaling_supp}
\end{figure*}

We evaluate the effect of the number of sampling steps on relighting quality in \cref{fig:inference_scaling_supp}. We report results using 1, 5, 10, 20, and 50 steps on the synthetic evaluation set described in \cref{subsec:quantitative_eval}. Each step requires two number-of-function-evaluations (NFE) due to classifier-free guidance. Timing is measured on a single A100 80GB GPU and reflects end-to-end inference time, including pre- and post-processing.

Both LPIPS and SSIM improve with more sampling steps, while PSNR exhibits minor fluctuations but follows an overall upward trend. Qualitatively, plausible lighting emerges even at very few steps, though with visible artifacts (e.g., residual noise patterns at 1 step in \cref{fig:inference_scaling_supp}(ii), and overly smooth appearance of the pumpkin at low step counts in \cref{fig:inference_scaling_supp}(iii)) that diminish as the step count increases. This suggests that distillation techniques \cite{consistency_models, dmd} may be a promising direction for accelerating inference in future work.

\section{Additional Qualitative Results}

\paragraph{Additional spatial lighting results}

\begin{figure*}[!htbp]
    \centering
    \includegraphics[width=\linewidth]{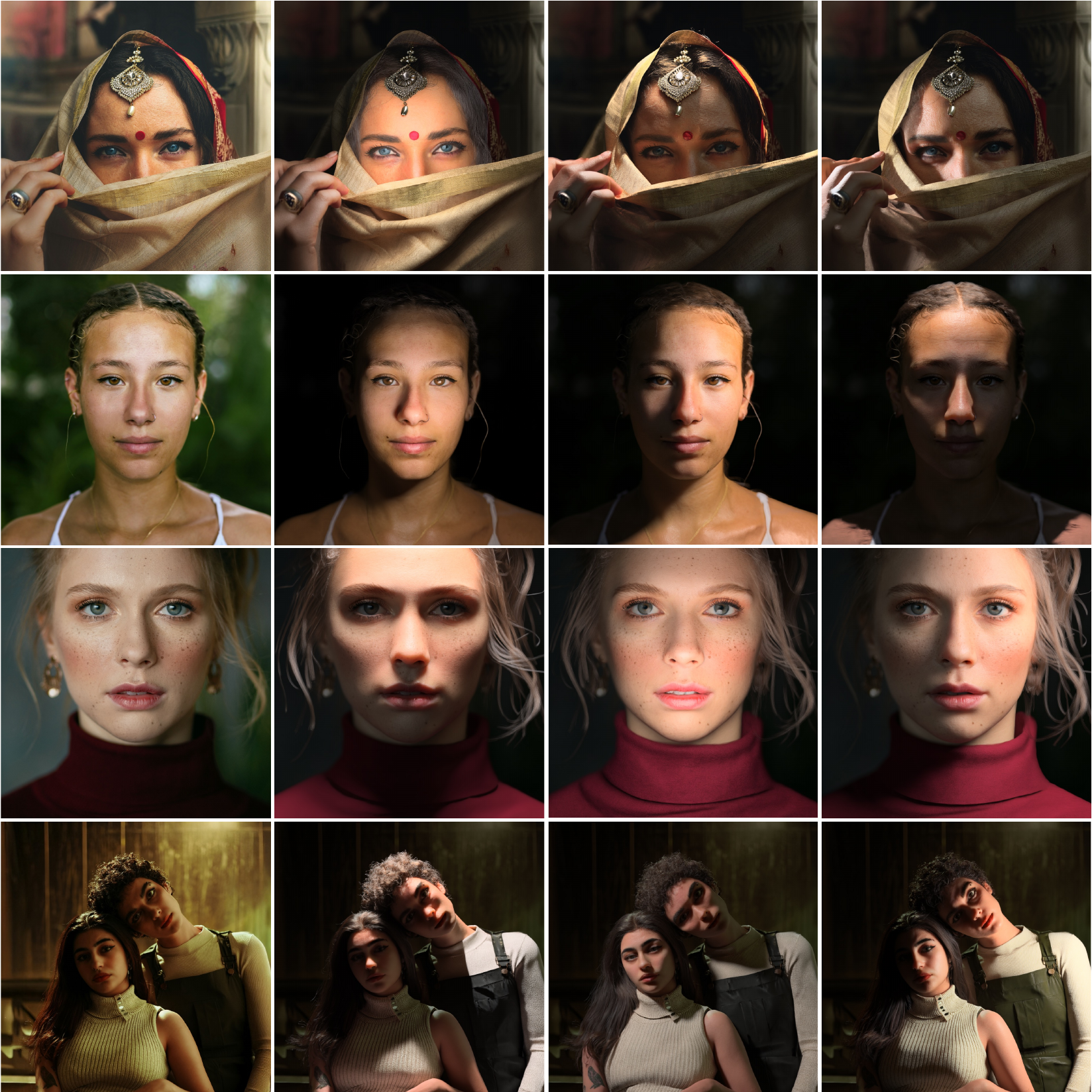}
    \caption{\textit{Spatial lighting for portraits}: For each input, we show three relit outputs with lights inserted at different 3D locations. The results demonstrate high-quality portrait relighting and complex light–geometry interactions, including challenging cases such as the veiled subject.}
    \label{fig:additional_portrait_results}
\end{figure*}

\cref{fig:additional_portrait_results} presents additional examples of inserting virtual lights at different 3D locations, with emphasis on portrait relighting applications. Our method produces high-quality results that preserve subject identity while robustly handling complex light–geometry interactions.

\paragraph{Spatial lighting under extreme viewpoints} 

\begin{figure*}[!htbp]
    \centering
    \includegraphics[width=\linewidth]{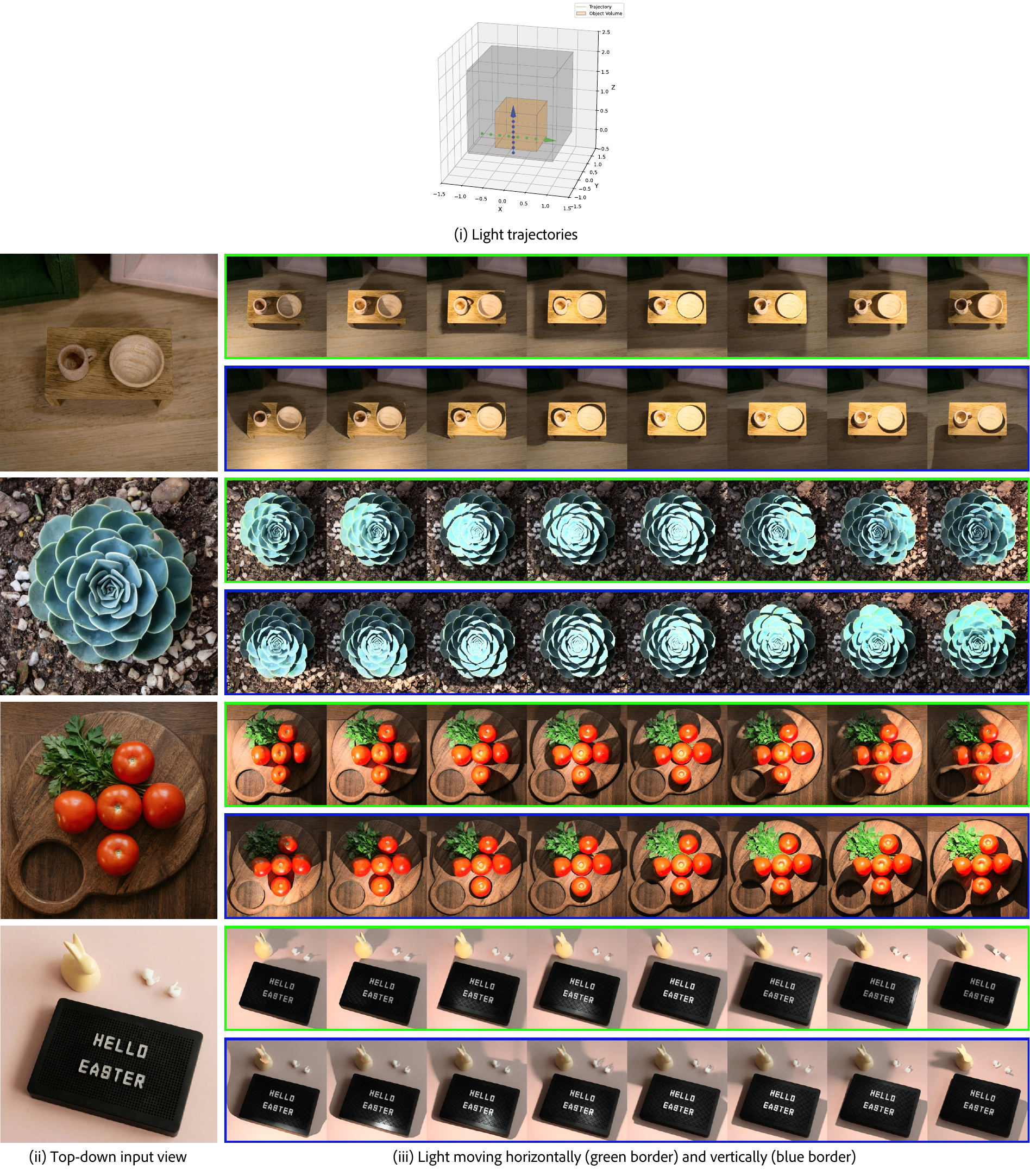}
    \caption{\textit{Extreme camera viewpoints.} We test on top-down views (left) moving the point light along two trajectories (top). The model follows both motions, left $\rightarrow$ right (green border) and down $\rightarrow$ up (blue border), with coherent shading and shadows.}
    \label{fig:top_down_cam}
\end{figure*}

\cref{fig:top_down_cam} shows results on images captured from top-down camera viewpoints. These viewpoints are not explicitly seen during training. We evaluate two virtual light trajectories: (a) left-to-right and (b) bottom-to-top, while the light is placed above the objects in both cases. Despite this challenging configuration, our method produces consistent shading and shadows. We attribute this behavior to our camera-agnostic lighting representation and the heavy augmentation of camera viewpoints during training, where we randomize the camera pitch in the range $\left[-30^\circ, 0^\circ\right]$.

\paragraph{Additional comparisons with Careaga et al.}

\begin{figure*}
    \centering
    \includegraphics[width=\linewidth]{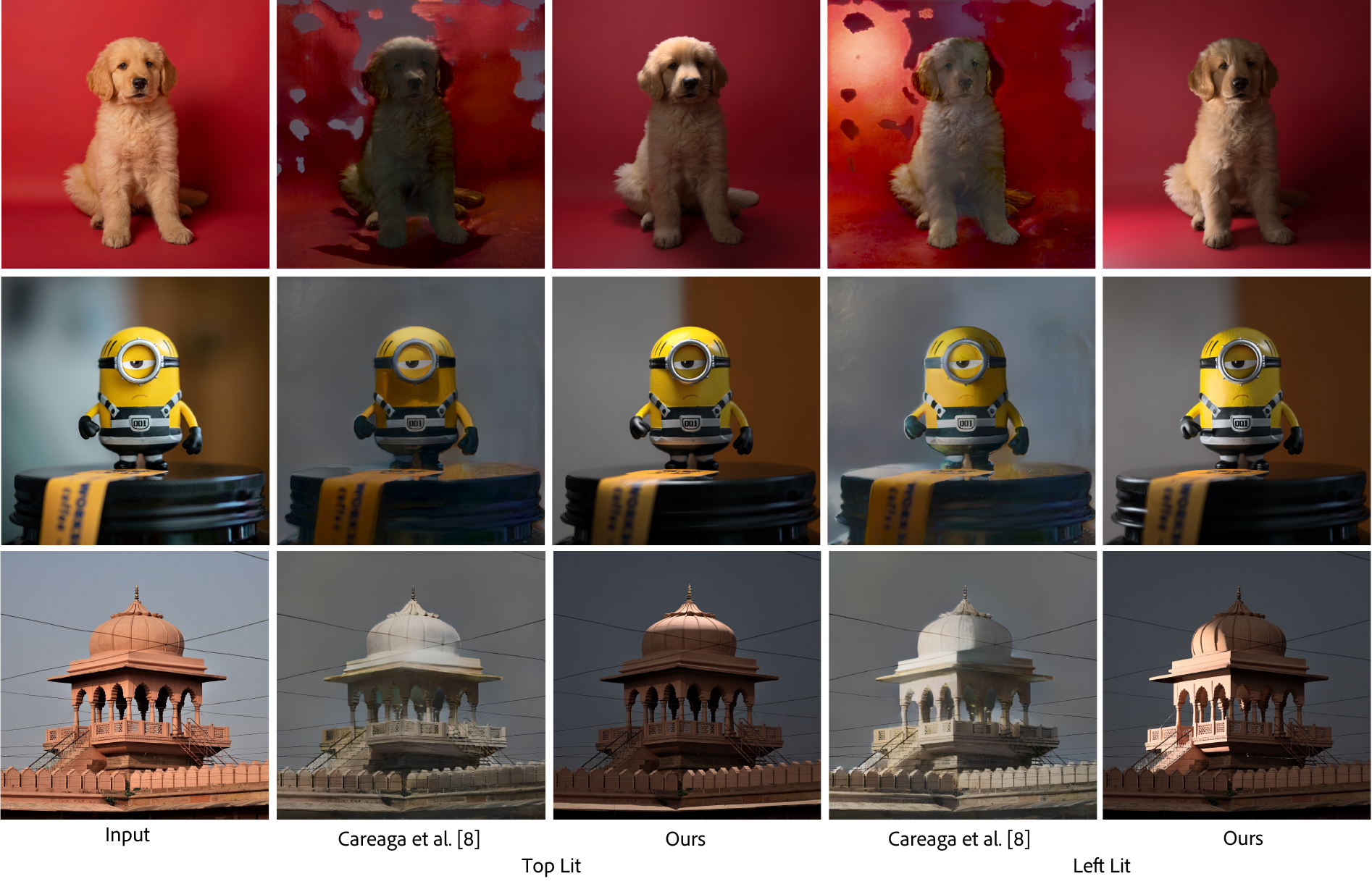}
    \caption{\textit{Additional comparison with Careaga et al. \cite{physically_controllable_relighting_sig_25}} We show the input image and results for two lighting directions (top-lit and left-lit). In the first row, our method produces plausible relighting on fur. In the second row, it better preserves the material appearance of the input. In the last row, it relights the building while maintaining its albedo.}
    \label{fig:supp_comparison_careaga}
\end{figure*}

\cref{fig:supp_comparison_careaga} provides further qualitative comparisons under additional lighting directions. Our method maintains consistent edits across diverse scenes, while Careaga et al. occasionally introduce unintended appearance changes—such as albedo shifts—likely stemming from intrinsic decomposition errors in their pipeline.

\begin{figure*}
    \centering
    \includegraphics[width=\linewidth]{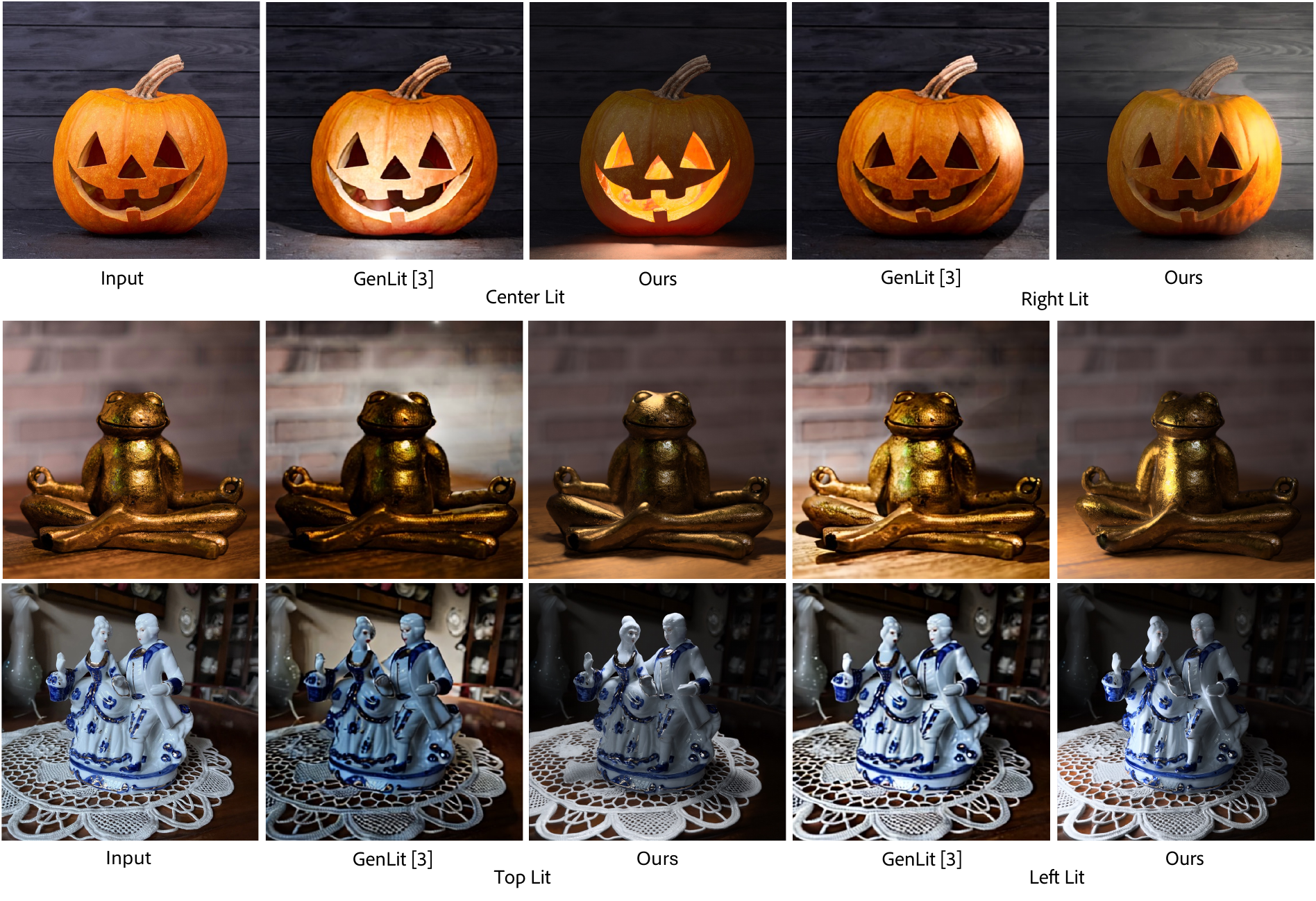}
    \caption{Additional comparison with GenLit \cite{genlit_sig_asia_shrisha_2025} Across these additional examples, our method produces more consistent light placement, while GenLit often exhibits drift in the intended light position.}
    \label{fig:supp_comparison_genlit}
\end{figure*}

\paragraph{Additional comparisons with GenLit}
\cref{fig:supp_comparison_genlit} shows further examples comparing our approach with GenLit. Across these cases, our method has stable light placement and coherent shading, whereas GenLit’s outputs often exhibit drift in the intended light position and less consistent illumination effects.

\paragraph{Ambient Lighting Control}

We provide additional qualitative results demonstrating the range of ambient–illumination edits supported by our model. 

\begin{figure*}[!ht]
    \centering
    \includegraphics[width=\textwidth]{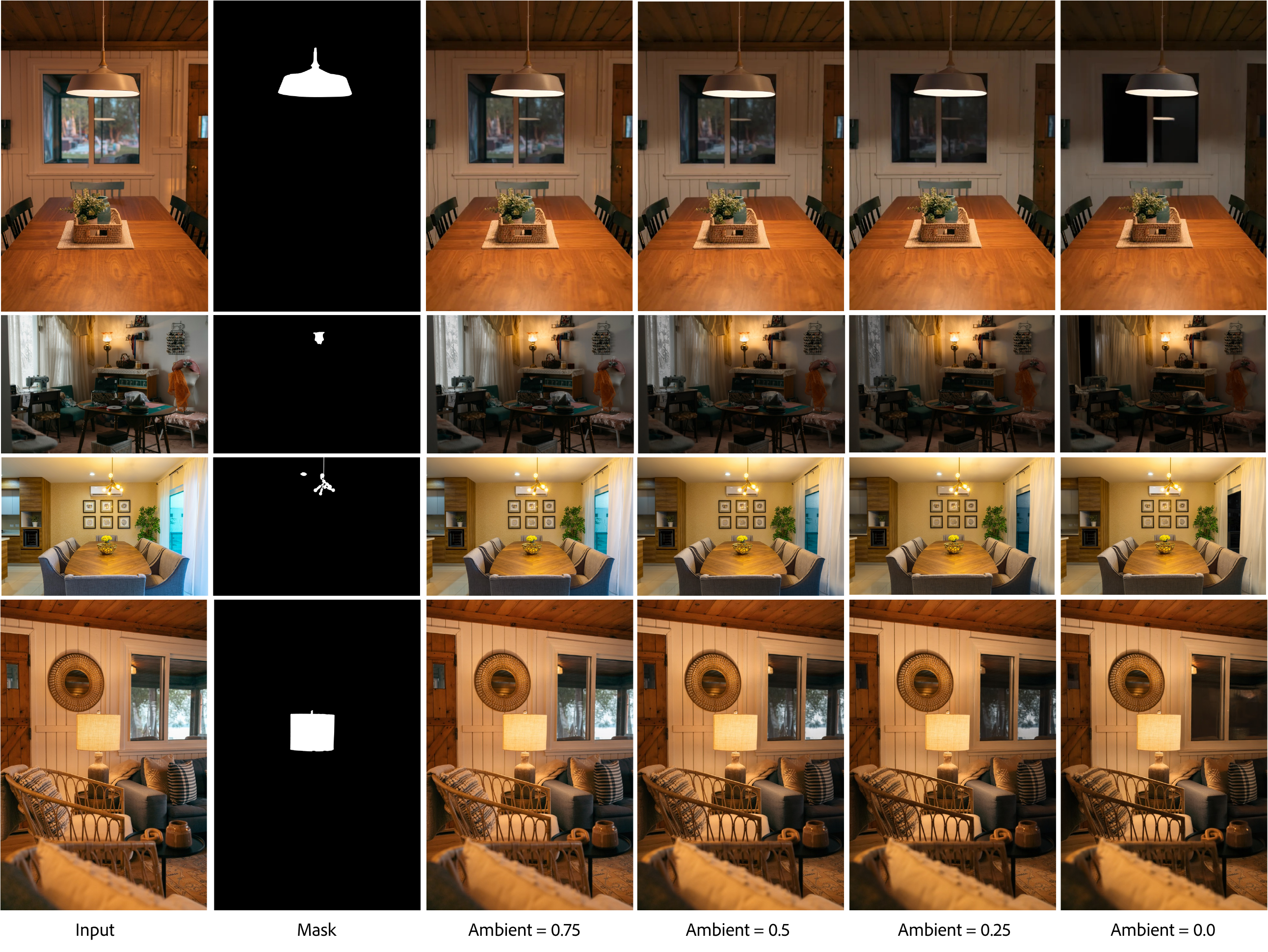}
    \caption{\textit{Continuous ambient–intensity control.} Each row shows an input image and its light-fixture mask. All remaining illumination is treated as ambient and is progressively reduced by our method. In the first row, for instance, the ambient light from the window fades while the masked light source’s reflection in the glass remains unchanged, illustrating proper separation of ambient and the masked light-fixture.}
    \label{fig:ambient_scaling_1}
\end{figure*}

\cref{fig:ambient_scaling_1} shows continuous scaling of ambient intensity on a variety of examples where a mask is used to specify the light source to be preserved while the remaining light sources, considered ambient lighting, are gradually dimmed. 

\begin{figure*}[!htbp]
    \centering
     \includegraphics[width=\linewidth]{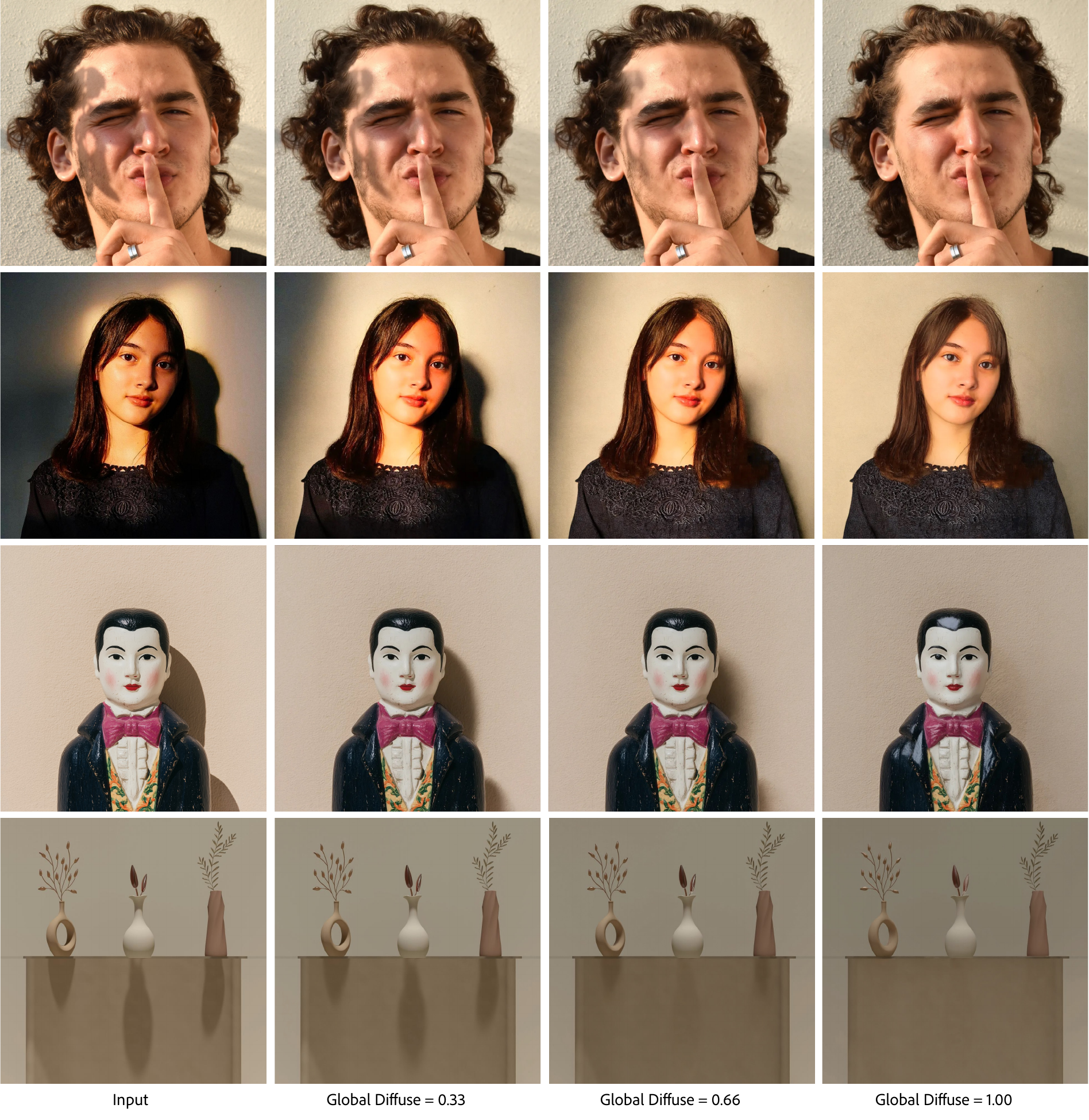}
    \caption{\textit{Ambient-light diffusion for shadow softness}. In each row, we show an input and three results with increasing global diffuse levels that increase shadow softness.}
    \label{fig:diffuse_softness_supp}
\end{figure*}

\begin{figure*}[!htbp]
    \centering
    \includegraphics[width=\linewidth]{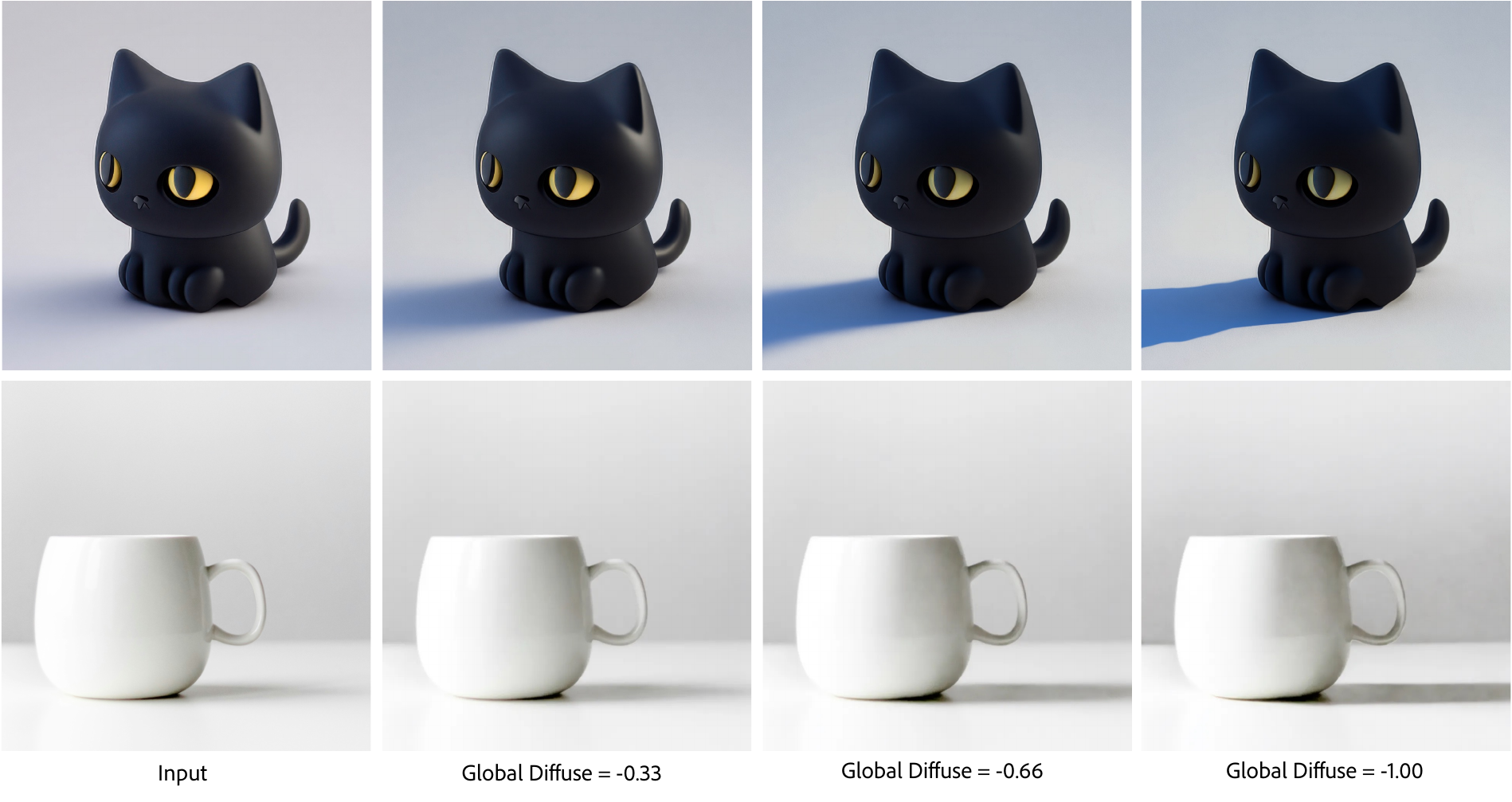}
    \caption{\textit{Ambient-light diffusion for shadow sharpening}. In each row, we show an input and three results with different global diffuse levels. By providing negative $d_g$ values, we can adjust the ambient light diffuse level to make shadows sharper.}
    \label{fig:diffuse_sharpness_supp}
\end{figure*}

Additionally, we demonstrate control over shadow softness using the diffuse spread parameter $d_g$. In \cref{fig:diffuse_softness_supp}, positive values of $d_g$ increase shadow softness by diffusing the light. Since our model is conditioned on the difference in spread parameters, $d_g$ can also be negative. As shown in \cref{fig:diffuse_sharpness_supp}, negative $d_g$ values reverse the light diffusion process, progressively sharpening shadows in the input image.

\paragraph{Visible-Light Fixture Intensity Control}

\begin{figure*}
    \centering
    \includegraphics[width=\linewidth]{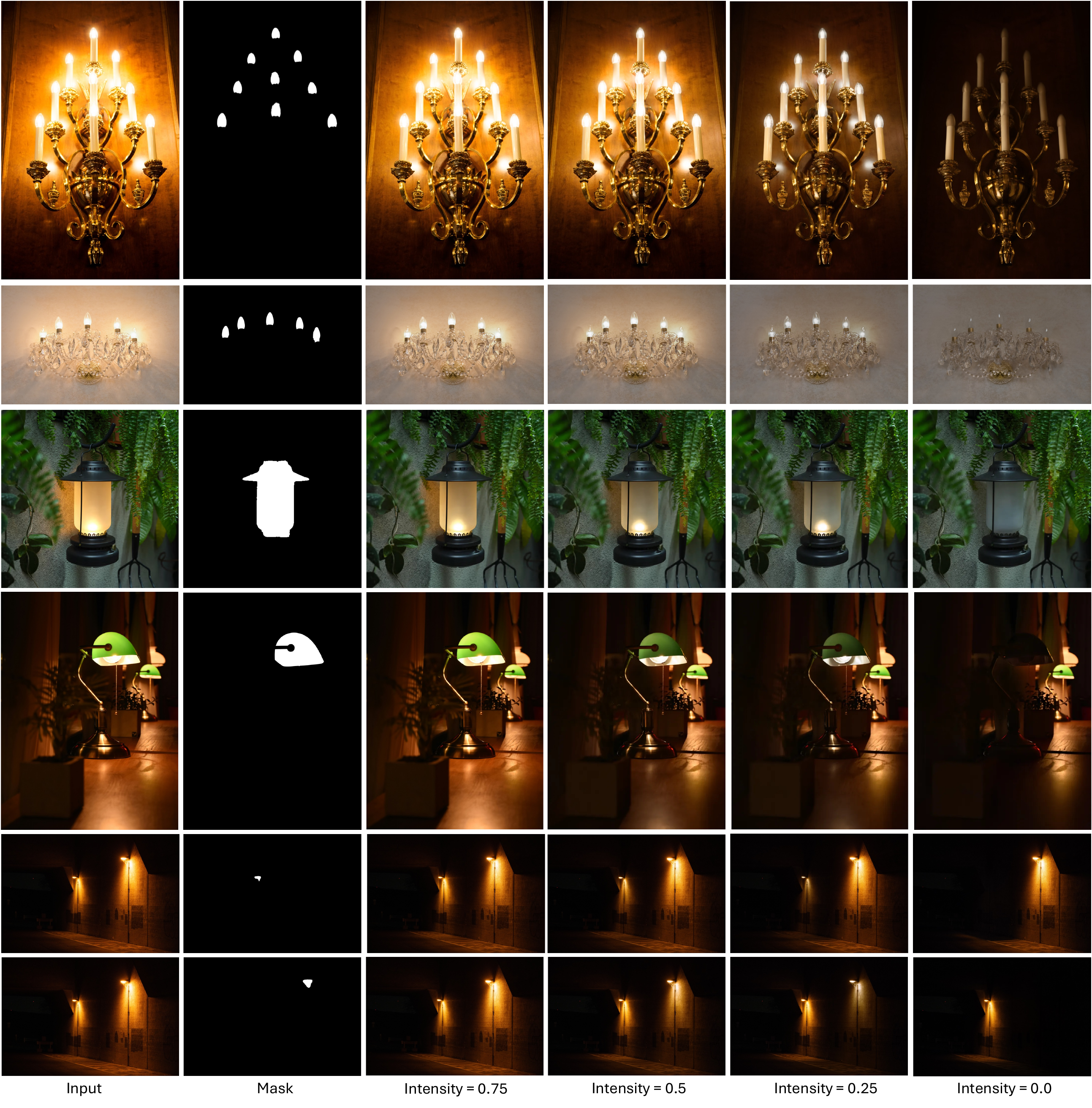}
    \caption{\textit{Light-intensity control to gradually turn off lights}. Our method gradually turns off light in input images, provided a mask and different intensity levels, even generalizing to complicated light fixtures such as chandeliers, different types of lamps and streetlights. In the last two rows we show how the mask can be used to localize lighting edits, turning off street lights one at a time.}
    \label{fig:light_turn_off_supp}
\end{figure*}

\begin{figure*}
    \centering
    \includegraphics[width=\linewidth]{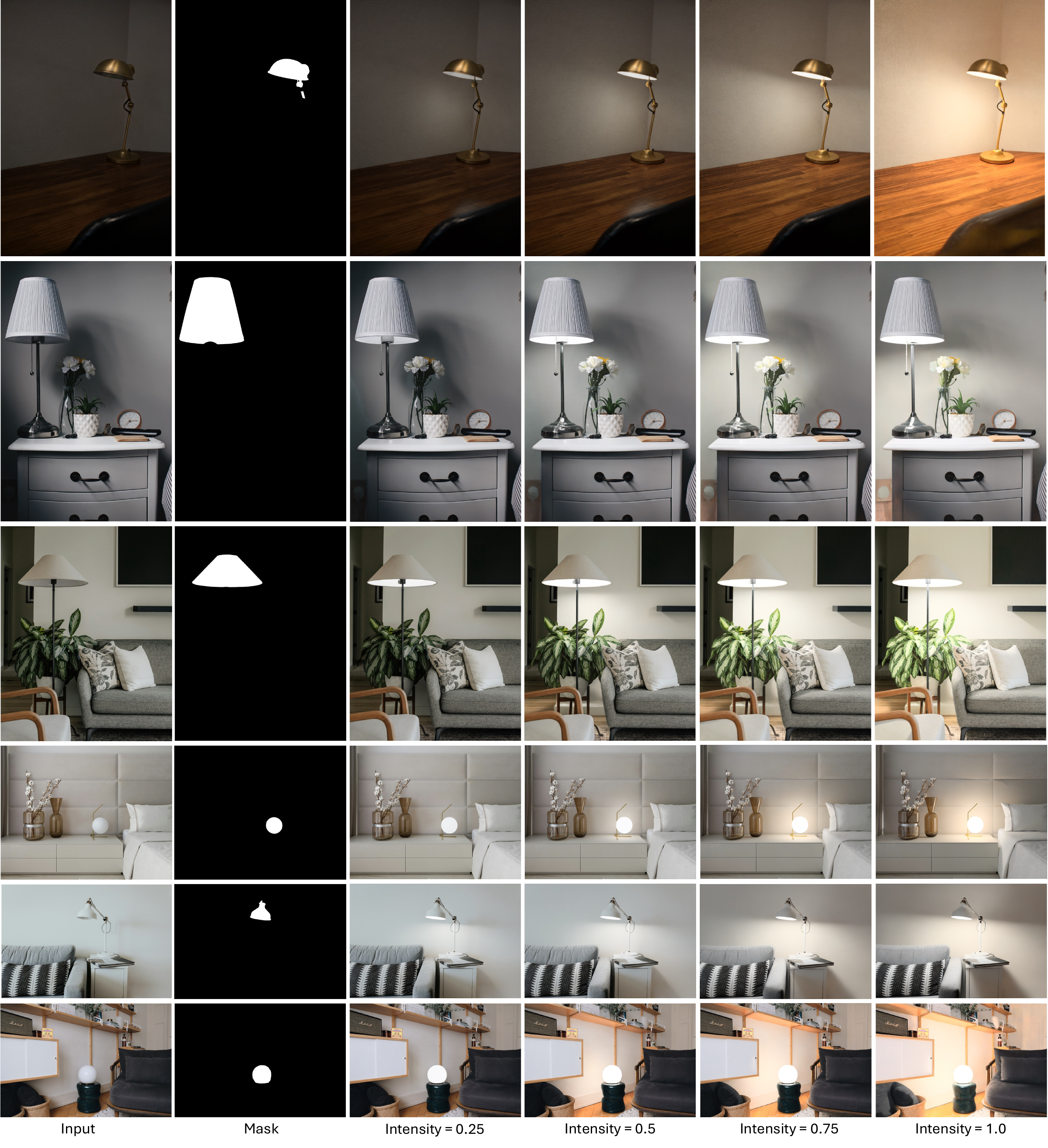}
    \caption{\textit{Light-intensity control for turning lights on}: Given a light-visibility mask, our method increases intensity with realistic light falloffs, i.e., the illuminated portions on nearby walls retain plausible boundary based on light-fixture shape.}
    \label{fig:light_turn_on_supp}
\end{figure*}

In \cref{fig:light_turn_off_supp} and \cref{fig:light_turn_on_supp}, we demonstrate fine-grained control over visible light-fixture intensities. 

In \cref{fig:light_turn_off_supp}, given an input image and a mask that localizes the visible light source, our method gradually dims the fixture, handling diverse cases including chandeliers. 

In \cref{fig:light_turn_on_supp}, we show the reverse process, where an initially off light is progressively turned on. As intensity increases, the illumination spread follows the geometry of the emitter i.e., the lamp shades and produces plausible, directionally consistent shadows throughout the scene. 

\paragraph{Outdoor Visible-Light Fixture Results} 

\begin{figure*}
    \centering
    \includegraphics[width=\linewidth]{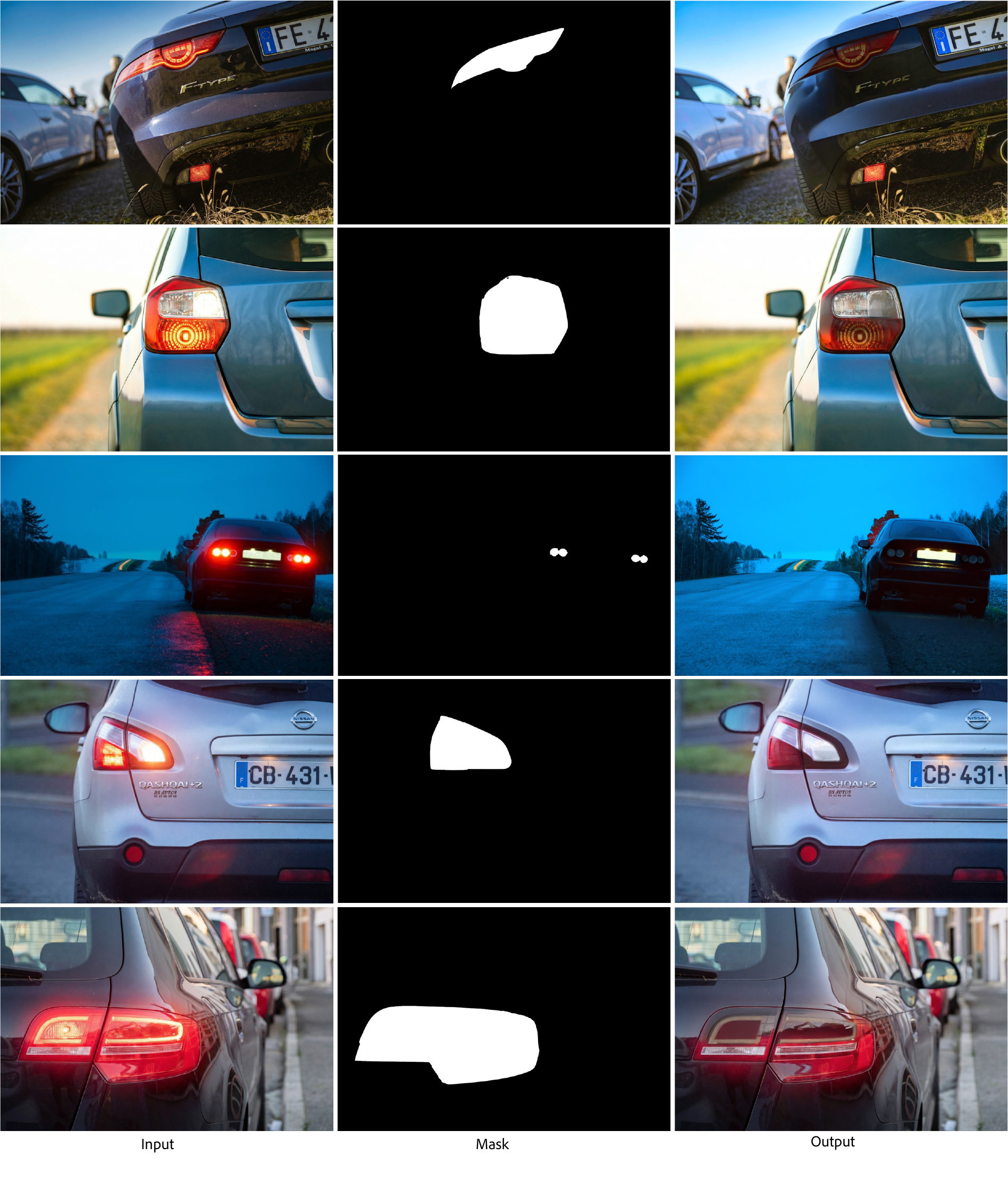}
    \caption{\textit{Outdoor light disentanglement.} Turning off a car’s backlight suppresses only the light, without darkening the environment. Our pipeline renders localized light edits under diverse environment maps, promoting this separation.}
    \label{fig:light_outdoor_headlight}
\end{figure*}

\cref{fig:light_outdoor_headlight} shows examples of localized light editing in outdoor scenes. In these results, turning off a car’s headlight suppresses only the light emitted by the fixture, while the surrounding illumination from the environment (e.g., sunlight) remains unchanged. Although our method is not explicitly trained on outdoor scenes, our synthetic dataset contains only indoor ones, we render training samples under a variety of environment maps, including outdoor HDRIs. Exposure to such lighting conditions may help the model distinguish localized visible light sources from global environment illumination, enabling plausible behavior in these outdoor examples.

\paragraph{Additional results on VisibleFixture-60}

\begin{figure*}[!htbp]
    \centering
    \includegraphics[width=0.95\linewidth]{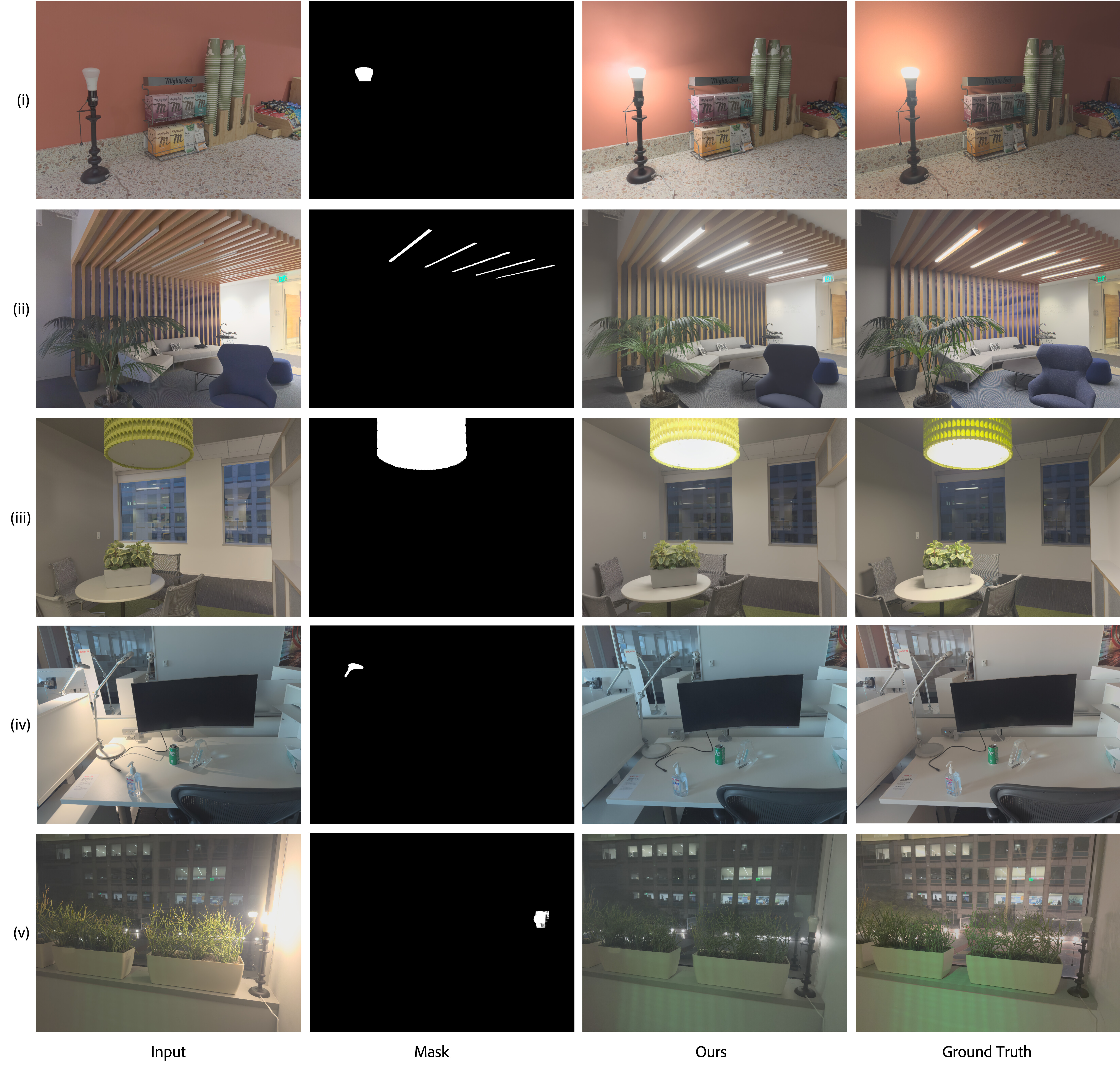}
    \caption{We present additional results on our \textit{VisibleFixture-60} test set with available ground truth. Rows (i-iii) show examples where our method turns lights on while rows (iv-v) show examples where our method turns lights off.}
    \label{fig:supp_visible_fixtures_60}
\end{figure*}

\cref{fig:supp_visible_fixtures_60} shows additional results from our VisibleFixture-60 test set, which provides paired captures with visible light sources toggled on and off. Our method turns lights on with illumination that closely matches the reference (row (i)), handles complex disjoint masks involving multiple fixtures (row (ii)), and produces shadows consistent with the captured scene (rows (ii–iii)). Rows (iv–v) show the reverse case: when lights are turned off, associated effects such as shadows (row (iv)) and reflections in the glass pane (row (v)) correctly disappear. 

\paragraph{Qualitative comparisons with LightLab}

\begin{figure*}[!htbp]
    \centering
    \includegraphics[width=\linewidth]{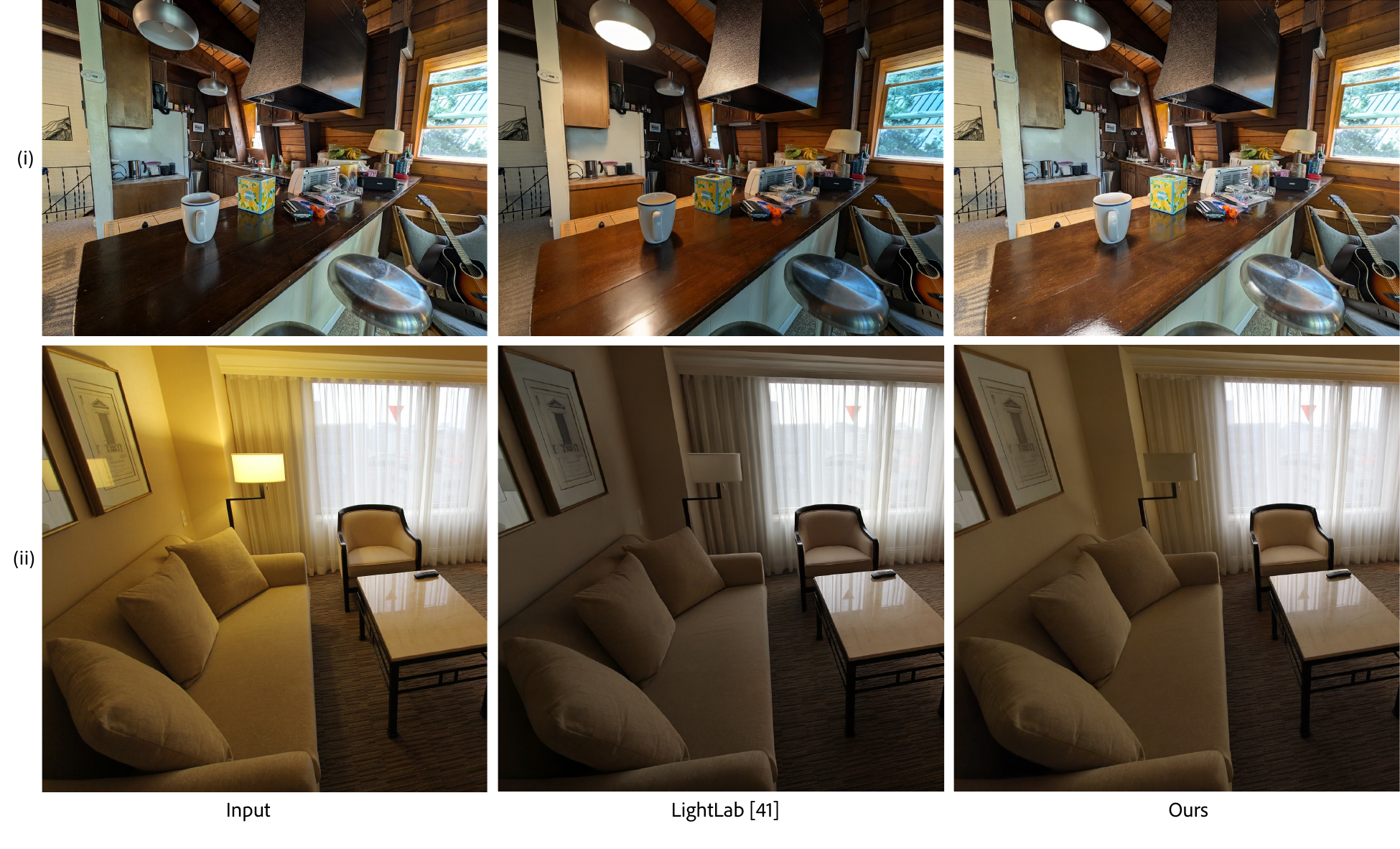}
    \caption{Comparison with LightLab \cite{light_lab_sig_25}: Using results provided by the authors, we qualitatively compare our method---reproducing photorealistic effects such as mug shadows and light reflecting from tabletop in (i), and the lamp turning off in (ii).}
    \label{fig:comparison_lightlab}
\end{figure*}

LightLab \cite{light_lab_sig_25} focuses on toggling visible light sources in real indoor scenes. Using author provided test-cases, we perform qualitative comparisons in \cref{fig:comparison_lightlab}. Our method reproduces fine-grained effects such as mug shadows and scene reflections (i), and turns lights off while retaining realistic ambient lighting (ii).

\section{Independent control of multiple lights}

Our representation extends trivially to multiple lights by repeating a compact per-light token block $(\mathbf{p}, \mathbf{c}, \lambda, d)$. Each block parameterizes a single light via its 3D position $\mathbf{p}$, color $\mathbf{c}$, intensity $\lambda$, and diffuse level $d$. We train a model variant supporting up to three lights. During training, we sample $k \in \{1,2,3\}$ active lights. For inactive lights, we set their parameters to $-1$.

Extending the single-light formulation in \cref{subsec:synthetic_data}, we construct supervision by summing the contributions of individual lights. Let $O_i$ denote the render corresponding to the $i$-th light at position $\mathbf{p}_i$ with sampled parameters $(\mathbf{c}_i, \lambda_i)$, and let $I$ denote the ambient render. The relit image is given by
\begin{equation}
I_r = \mathbf{T}\!\left(a\,I + \sum_{i=1}^{k} \lambda_i\,\mathbf{c}_i\,O_i \right),
\end{equation}
where $a \in [0,1]$ is the ambient scale and $\mathbf{T}(\cdot)$ is the tone-mapping operator. This construction mirrors the single-light case, with the controllable illumination formed by summing per-light contributions.

Since our lighting representation is compact, repeating the per-light token block results in only a modest increase in sequence length; for up to three lights, this incurs negligible additional training and inference cost.

\begin{figure*}[t]
    \centering
    \begin{subfigure}{\linewidth}
        \centering
        \includegraphics[width=\linewidth]{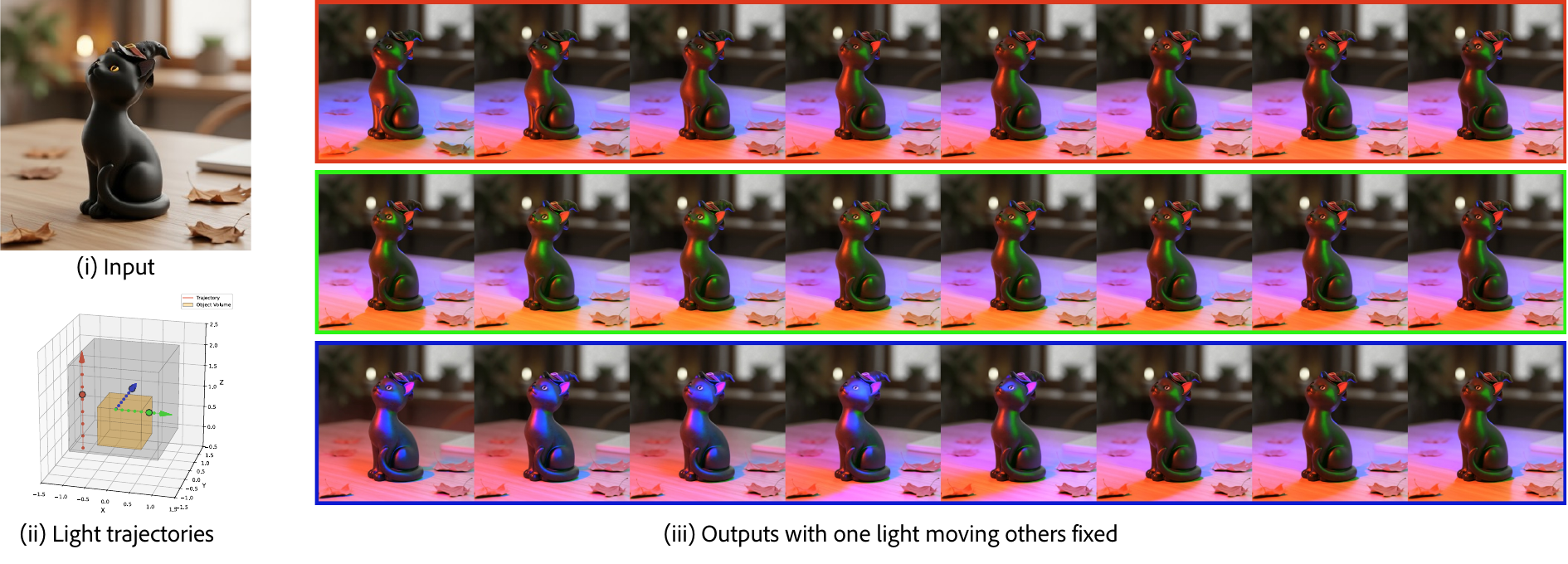}
        \caption{\textit{Independent multi-light control}. We show the input (top left), the 3D trajectories of three light sources (bottom left), and relighting results obtained by sweeping one light while keeping the other two fixed (right): Top row shows red light moving from bottom to top, middle row shows green light moving from center to right and bottom row shows blue light moving from front to back.}
            \label{fig:cat_multi_light}
    \end{subfigure}

    \vspace{0.5em}

    \begin{subfigure}{\linewidth}
        \centering
        \includegraphics[width=\linewidth]{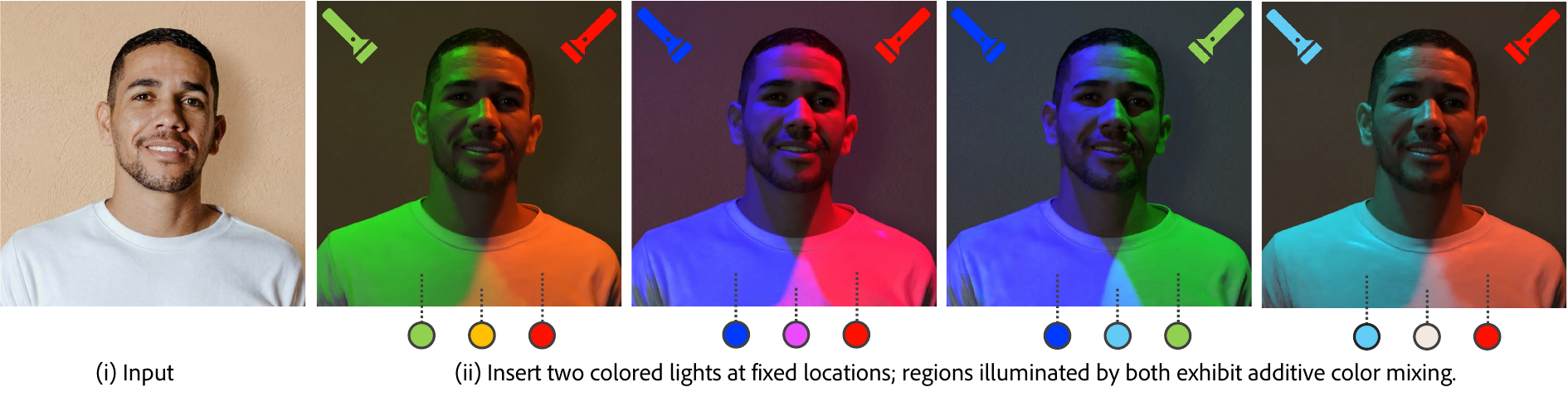}
        \caption{\textit{Color mixing under multiple lights}. We show the input (left) and relit outputs (right) under two fixed light sources placed at the top-left and top-right of the scene. The model produces plausible color mixing and colored shadows, following expected additive behavior across both primary (first three examples) and complementary color combinations (i.e., red + cyan = white).}
        \label{fig:person_multi_light}
    \end{subfigure}

    \caption{Our compact light representation can easily be extended to support simultaneous editing of multiple light sources with different attributes (location, intensity, color, diffuse level) within a single inference pass.}
    \label{fig:multi_light_results}
\end{figure*}

\cref{fig:multi_light_results} shows qualitative results with two and three lights. The model produces consistent shading and shadows, with independent light control \cref{fig:cat_multi_light} and plausible color mixing along shadow boundaries \cref{fig:person_multi_light}.

\section{Discussion on Video Relighting}

Video relighting is a natural extension of our work, reflected by the recent surge of interest \cite{fang2025relightvidtemporalconsistentdiffusionmodel, liu2025uniLumos, diffusion_renderer_cvpr_25}. However, extending our lighting representation to video raises interesting research questions, which we discuss here with a focus on spatial lighting control.

The main challenge arises from the fact that the operational space of our spatial lighting representation (i.e., the 3D coordinates exposed to the user) is defined relative to the camera rather than in a canonical 3D world space, raising questions about light placement and persistence as objects/camera move within the video. In the simplest setting, without a driving video, with the goal of animating a relit image, an image-to-video model \cite{wan2025wanopenadvancedlargescale} may be used to propagate lighting edits. Similarly, in settings with a driving video but limited object motion and a static camera (e.g., facial performance capture \cite{he2024diffrelight}), our camera-agnostic lighting representation may already be sufficient, provided the model is trained on paired relighting data with the appropriate motion characteristics.

In the general case, where both camera and scene objects move, it remains unclear how a lighting edit specified in the first frame should persist over time. This raises an open question in how to formulate the learning problem---whether via explicit pose tracking, per-frame light tokens, or implicit inference from video. We view this as an interesting direction for future work.

Finally, future work can address autoregressive generation for faster response times. Recent advances \cite{huang2025self, slow_to_fast_tianwei_yin_cvpr_2025} in efficient autoregressive video generation suggest a plausible route toward interactive video relighting.

\section{Limitations}

\begin{figure*}[!htbp]
    \centering
    \includegraphics[width=\linewidth]{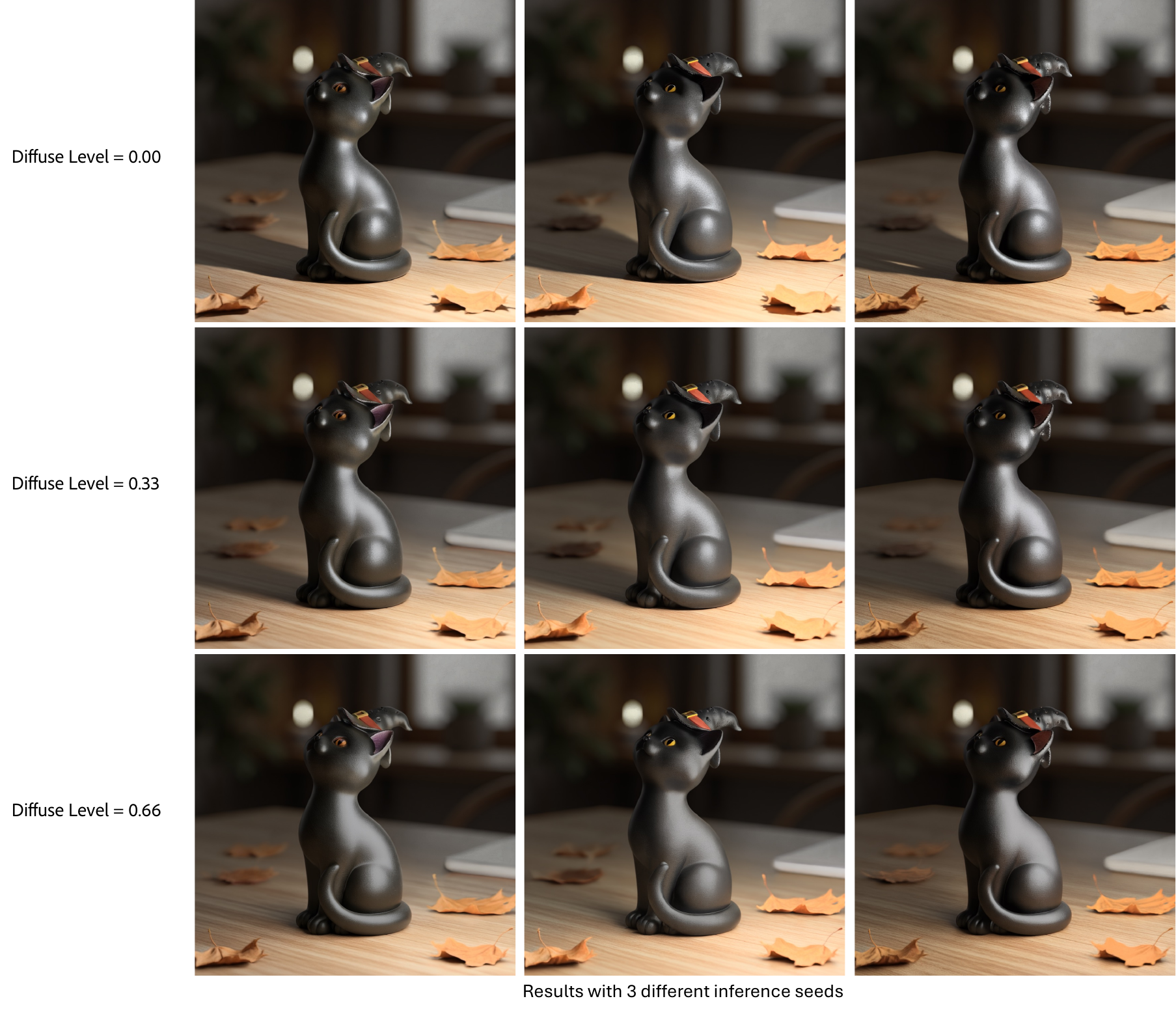}
    \caption{\textit{Seed-dependent variation across diffuse levels.} We insert a virtual light at the same 3D location and vary its diffuse level across the three rows (low to high). Each row shows three outputs generated with different inference seeds. At low diffuse levels, shadows become sharper and exhibit minor seed-dependent shifts. As the diffuse level increases, results become naturally more consistent across seeds. All outputs remain plausible and preserve the intended lighting edit, illustrating that seed-dependent variation is limited to subtle differences.}

    \label{fig:limitations}
\end{figure*}

While our tokenized representation enables intuitive control, several limitations remain. First, our method relies on a large DiT architecture, making real-time interaction for applications requiring immediate feedback currently challenging. Second, the diffusion-based formulation introduces sampling stochasticity—different random seeds produce visually similar but non-identical results (\cref{fig:limitations})---most visible at the low-diffuse settings with sharper shadows---which may be problematic for workflows requiring deterministic outputs. Finally, our method generalizes better to indoor scenes than outdoor environments, reflecting a training data bias: our dataset contains no outdoor synthetic scenes. Future work could address these limitations through model distillation for faster inference and training with more diverse outdoor data.





\end{document}